\documentclass[numbib]{imaiai}
\usepackage{amsmath,amsfonts,algorithm,algorithmic,subfig,url,comment,multirow} 
\usepackage{graphicx}

\def\x{{\mathbf x}}

\def\RN{ \mathbb{R} }                               

\newcommand{\bx}{\boldsymbol{x}}

\newcommand{\by}{\boldsymbol{y}}

\newcommand{\EE}{{\mathbb E}}
\newcommand{\PP}{{\mathbb P}}

\newcommand{\ba}{\boldsymbol{a}}

\newcommand{\bA}{\boldsymbol{A}}
\newcommand{\bB}{\boldsymbol{B}}

\newcommand{\bI}{\boldsymbol{I}}

\newcommand{\A}{\mathcal{A}}
\newcommand{\B}{\mathcal{B}}
\newcommand{\C}{\mathcal{C}}
\newcommand{\F}{\mathcal{F}}

\newcommand{\bGa}{\boldsymbol{\Gamma}}

\newcommand{\eg}{{\em e.g., }}
\DeclareMathOperator*{\argmin}{arg\,min}

\begin{document}

\title{Semi-Supervised Single- and Multi-Domain Regression with Multi-Domain Training}

\shorttitle{Semi-Supervised Multi-Domain Regression} 
\shortauthorlist{Michaeli, Eldar and Sapiro} 

\author{{
\sc Tomer Michaeli}$^*$,\\[2pt]
Technion--Israel Institute of Technology\\
$^*${\email{tomermic@tx.technion.ac.il}}\thanks{This paper was submitted for possible publication in \it{Information and Inference: A Journal of the IMA}.}\\[2pt]
{\sc Yonina C.\ Eldar}\\[2pt]
Technion--Israel Institute of Technology\\
{yonina@ee.technion.ac.il}\\[6pt]
{\sc and}\\[6pt]
{\sc Guillermo Sapiro} \\[2pt]
University of Minnesota\\\
{guille@umn.edu}}

\maketitle

\begin{abstract}
{We address the problems of multi-domain and single-domain regression based on distinct and unpaired labeled training sets for each of the domains and a large unlabeled training set from all domains. We formulate these problems as a Bayesian estimation with partial knowledge of statistical relations. We propose a worst-case design strategy and study the resulting estimators. Our analysis explicitly accounts for the cardinality of the labeled sets and includes the special cases in which one of the labeled sets is very large or, in the other extreme, completely missing. We demonstrate our estimators in the context of removing expressions from facial images and in the context of audio-visual word recognition, and provide comparisons to several recently proposed multi-modal learning algorithms.}
{Bayesian estimation, partial knowledge, multi and single domain regression, learning, hidden relationships, Bayesian networks, minimum mean squared error.}
\end{abstract}

\section{Introduction}
\label{sec:introduction}

There are many applications in which one can access data from multiple domains in order to perform a task. For example, word recognition can greatly benefit from the availability of joint audio-visual measurements \citep{NKKNLN10}. Person recognition and verification can be performed much more accurately by fusing information from several modalities such as facial images, iris scans, voice recordings, and handwritings.

A major difficulty in fusing multiple sources is that one can often access only distinct labeled training sets for the different domains and does not have paired labeled examples from all domains. Suppose, for instance, we wish to perform audio-visual gender recognition. There are numerous existing data-sets of labeled voice recordings as well as labeled data-sets of facial images. However, there are only a few jointly labeled audio-visual data-sets, with a limited number of different subjects each. Thus, although it is straight forward to train a classifier based on audio or image data alone, it is not clear how to best fuse the two modalities, in particular when they are unpaired. While paired multi-domain labeled examples are typically scarce, paired unlabeled examples are often abundant. For instance, enormous amounts of speaker video sequences (together with audio) can be easily collected. These videos, though, often do not come with labels. Nonetheless, they can be used to unveil the statistical relations between audio and video. An important question is how to best fuse audio- and image-based predictors, given these relations.

An even more interesting and practical question is whether the availability of multiple data sources can aid a machine learning algorithm during training, even if not all are measured during testing. For example, suppose we want to predict the age of a person based on an audio recording of him/her. Assume we have a labeled audio training set, a labeled image training set, and a large amount of unlabeled audio-visual examples. Can the visual examples help construct a predictor, which is solely based on audio?

In this paper we address the problem of multi-domain as well as single-domain regression based on distinct (unpaired) labeled training sets for each of the domains and an unlabeled multi-domain training set. Specifically, focusing on two domains for simplicity, we consider the situation in which we have at our disposal a very large unlabeled training set $\{\bx_1^i,\bx_2^i\}$ as well as two labeled sets $\{\bx_1^i,\by^i\}$ and $\{\bx_2^i,\by^i\}$. Using this multi-domain training data, we treat the problems of designing a predictor of $\by$ based on $(\bx_1,\bx_2)$ (multi-domain regression) and a predictor of $\by$ based on $\bx_1$ alone (single-domain regression). Our analysis is general in that it explicitly accounts for the cardinality of the labeled sets. In particular, it includes the special cases in which one or both labeled sets are very large as well as the cases in which one of the labeled sets is completely missing.

Several problems of similar nature have been treated in the literature. Perhaps the most widely studied of these is \emph{multi-view learning} \citep{BM98} in general and multi-view regression \citep{KF07} in particular. These techniques make use of a large training set of data from multiple domains (views), containing only a few labeled examples. It has been shown that if the views tend to agree in some sense, then the unlabeled examples are useful in constructing a single-view estimator \citep{BM98,KF07}. In our setting, however, we do not observe even a single multi-domain labeled example $\{\bx_1^i,\bx_2^i, \by^i\}$ and also make no assumptions on the underlying distribution. A multi-view framework for distinct labeled training sets, recently proposed in \cite{AUG09}, assumes the availability of a mapping function which can generate a good estimate of the unobserved view from the observed one. In our setting, we do not assume that such a mapping is known or even exists. These distinctions have profound implications. In particular, the lack of labeled multi-domain samples in our scenario implies that, even if our single-domain sets are infinite, we may only be able to deduce the joint distribution of $(\bx_1,\bx_2)$, of $(\bx_1,\by)$, and of $(\bx_2,\by)$. This, however, does not suffice, in general, to determine the conditional distribution of $\by$ given $(\bx_1,\bx_2)$, and therefore, for instance, the minimum mean-square error (MMSE) estimator $\rho(\bx_1,\bx_2)=\EE[Y|X_1=\bx_1,X_2=\bx_2]$ cannot be constructed.

Situations in which labeled samples $\{\bx_2^i,\by^i\}$ from a source domain are used to construct a predictor of $\by$ from a target domain $\bx_1$ fall under the category of \emph{transfer learning} \citep{PY10}. In some cases, unlabeled examples, as well as a few labeled examples $\{\bx_1^i,\by^i\}$ from the target domain are also available. Traditional transfer learning algorithms are suited for domains admitting a common feature representation. For example, the different domains may be images of an object taken from different views, in which case the extracted features are of the same type. Extension to different representations may be handled via the \emph{multiple-outlook learning} framework \citep{HM11}. Nevertheless, in both these settings paired unlabeled examples $\{\bx_1^i,\bx_2^i\}$ from the two domains are not accessible. In this sense, our setting allows learning via supervised-transfer of knowledge.

More related to our problem are the \emph{cross-modality} and \emph{shared-representation} learning scenarios recently studied in \cite{NKKNLN10} in the context of multi-modal learning. In both settings, unlabeled training data $\{\bx_1^i,\bx^i_2\}$ from multiple modalities, such as audio and video, are used to perform a \emph{feature learning} stage. In cross-modality learning, one constructs a predictor based on $\bx_1$ alone using a labeled training set $\{\bx_1^i,\by^i\}$. For example, we may want to build a classifier operating on audio features by observing labeled audio examples in addition to unlabeled audio-visual instances. In shared-representation learning, one constructs a predictor based on $\bx_1$ alone using a labeled training set $\{\bx_2^i,\by^i\}$. For instance, we may want to train an audio classifier by observing only labeled visual examples in addition to unlabeled audio-visual instances. Cross-modality regression was recently studied from a Bayesian estimation perspective in \cite{ME11}, in which a link to instrumental variable regression \citep{BT84} was also highlighted. As we show, both cross-modality and shared-representation learning are special cases of our approach, corresponding to the situation in which there are zero examples in one of the labeled sets.

In this paper we formulate regression from unpaired data sets as a Bayesian estimation problem with partial knowledge of statistical relations. Specifically, we assume that, for each domain, we can determine the predictor that minimizes the mean square error (MSE) among some class of estimators. This can be done using the labeled training examples from the associated domain. Furthermore, we assume that we can determine the joint probability distribution of the data from the two domains using the unlabeled examples. Now, every joint distribution of labels and (multi-domain) data which is consistent with this knowledge is considered valid. The performance of any estimator depends, of course, on the unknown distribution. Thus, our approach in this paper is to seek estimators whose worst-case MSE over the set of valid distributions is the smallest possible.

We show that the minimax problems we obtain have simple, yet nontrivial, closed form solutions which can be easily approximated from the available training examples. These expressions also provide insight into how data from multiple domains should be taken into account. In particular, we show that, from a worst-case standpoint, a domain with no labeled examples cannot help. Thus, it is impossible to perform cross-modality regression without making any assumptions on the underlying distributions. We illustrate our approach in the contexts of face normalization and audio-visual word recognition. In the former application, we demonstrate how an image of a smiling face can be converted into one with a neutral expression, without observing paired examples of neutral and smiling faces. In the latter setting, we show how spoken digits can be recognized from silent video (lipreading) when only labeled audio examples are available. We also show how they can be recognized from audio, when there is access only to labeled video examples. The experiments indicate that our approach is preferable to that of \cite{NKKNLN10}.

The remainder of this paper is organized as follows. In Section ~\ref{sec:ProbForm} we present the setting of interest in detail and discuss several special cases. We provide a mathematical formulation of our regression problems in Section~\ref{sec:EstTheoForm}. The minimax multi-domain and single-domain estimators are derived in sections~\ref{sec:MultiDomainRegression} and~\ref{sec:SingleDomainRegression}, respectively. Finally, experimental results are provided in Section~\ref{sec:experiments}.

\section{Problem Formulation}
\label{sec:ProbForm}
We denote random variables (RVs) by capital letters (\eg $X_1,X_2,Y$) and the values that they take by bold lower-case letters (\eg $\bx_1,\bx_2,\by$). The pseudo-inverse of a matrix $\bA$ is denoted by $\bA^\dag$. The second-order moment matrix of an RV $X$ is denoted by $\bGa_{XX}=\EE[XX^T]$, where $\EE[\cdot]$ is the mathematical expectation operator. Similarly, the cross second-order moment matrix of two RVs $X$ and $Y$ is denoted by $\bGa_{XY}=\EE[XY^T]$. The joint cumulative distribution function of the RVs $X$ and $Y$ is written $F_{XY}(\bx,\by)=\PP(X\leq\bx,Y\leq\by)$, where the inequalities are element-wise. By definition, the marginal distribution of $X$ is $F_{X}(\bx)=F_{XY}(\bx,\infty)$. In our setting, $Y$ is the quantity to be estimated, and $X_1$ and $X_2$ are two sets of measurements (features). The RVs $X_1$, $X_2$, and $Y$ take values in $\RN^{M_1}$, $\RN^{M_2}$, and $\RN^N$, respectively.

Our goal in this paper is to propose an estimation theoretic approach for solving certain regression problems in which several distinct training sets are available during training. More specifically, we assume we are given access to three possible data-sets as follows:
\begin{enumerate}
\setlength{\itemsep}{1pt}
\setlength{\parskip}{0pt}
\setlength{\parsep}{0pt}
\item labeled examples $\{(\bx_1^\ell,\by^\ell)\}_{\ell=1}^{L_1}$ from domain 1;
\item labeled examples $\{(\bx_2^\ell,\by^\ell)\}_{\ell=L_1+1}^{L_1+L_2}$ from domain 2;
\item paired unlabeled examples $\{(\bx_1^u,\bx_2^u)\}_{u=L_1+L_2+1}^{L_1+L_2+U}$.
\end{enumerate}
These training sets correspond to independent draws from the distributions $F_{X_1Y}$, $F_{X_2Y}$, and $F_{X_1X_2}$, respectively. Our focus is on situations in which $U$ is very large, so that the joint distribution $F_{X_1X_2}$ can be assumed known (or very well approximated, for example, by nonparametric methods). The cardinalities $L_1$ and $L_2$ of the labeled sets are arbitrary. In particular, one of them can be zero. In in this case no knowledge whatsoever is available regarding the statistical relation between $Y$ and the associated domain. On the other extreme, one (or both) of the labeled sets may be very large, in which case the associated single-domain MMSE estimator, say $\EE[Y|X_1]$, can be assumed known (or accurately approximated).

\begin{figure*}[t]\centering
\subfloat[]{\includegraphics[scale=1]{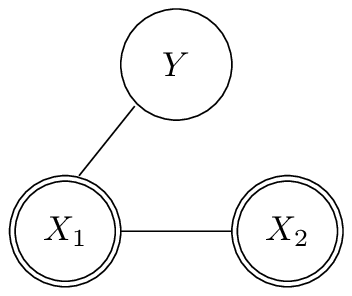}\label{fig:MultiDomReg_SingleDomTrainMany}}\hspace{0.1cm}
\subfloat[]{\includegraphics[scale=1]{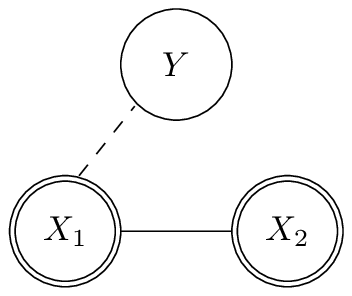}\label{fig:MultiDomReg_SingleDomTrainFew}}\hspace{0.1cm}
\subfloat[]{\includegraphics[scale=1]{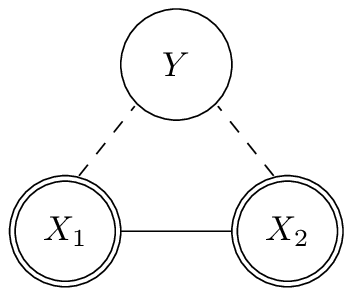}\label{fig:MultiDomReg_MultiDomTrainParam}}\hspace{0.1cm}
\subfloat[]{\includegraphics[scale=1]{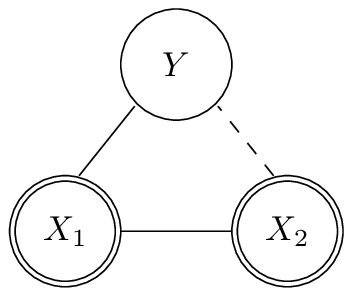}\label{fig:MultiDomReg_MultiDomTrainSemiParam}}
\caption{Multi-domain regression. (a),(b)~Single-domain training with many/few labeled examples (Section~\ref{sec:SingleDomTrainMultiDomReg}). (c)~Multi-domain training with few labeled examples (sections~\ref{sec:MultiDomLinReg} and \ref{sec:MultiDomParReg}). (d)~Multi-domain training with many unpaired labeled examples from one domain and few from the other domain (sections~\ref{sec:MultiDomPartialLinReg} and \ref{sec:MultiDomSemiParReg}).}
\label{fig:XYZmulti}
\end{figure*}

\begin{figure*}[t]\centering
\subfloat[]{\includegraphics[scale=1]{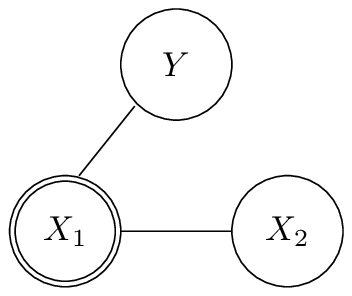}\label{fig:SingleDomReg_CrossModMany}}\hspace{0.1cm}
\subfloat[]{\includegraphics[scale=1]{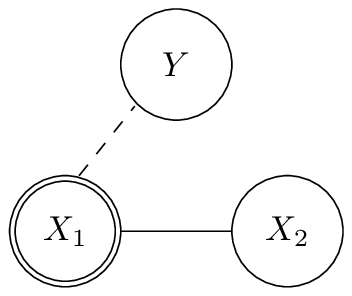}\label{fig:SingleDomReg_CrossModFew}}\hspace{0.1cm}
\subfloat[]{\includegraphics[scale=1]{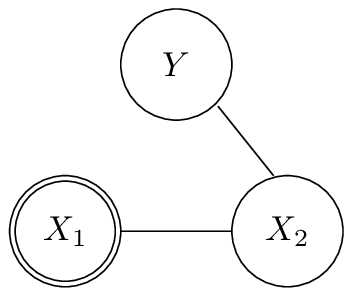}\label{fig:SingleDomReg_SharedRepMany}}\hspace{0.1cm}
\subfloat[]{\includegraphics[scale=1]{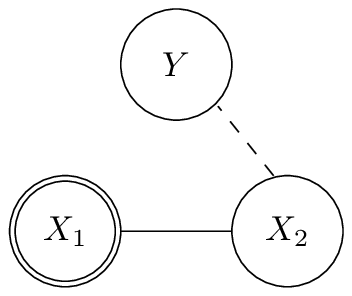}\label{fig:SingleDomReg_SharedRepFew}}\\
\subfloat[]{\includegraphics[scale=1]{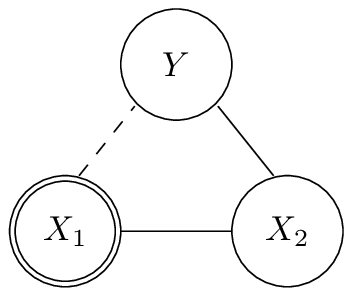}\label{fig:SingleDomReg_MultiDomTrainSemiPar}}\hspace{0.1cm}
\subfloat[]{\includegraphics[scale=1]{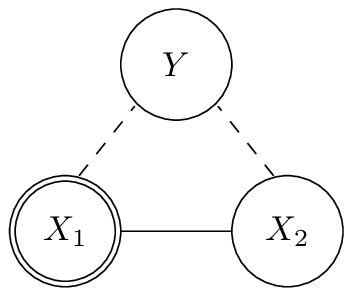}\label{fig:SingleDomReg_MultiDomTrainFew}}
\caption{Single-domain regression. (a),(b)~Cross-domain learning \citep{NKKNLN10} with many/few labeled examples (Section~\ref{sec:CrossDomain}). (c),(d)~Shared-representation regression \citep{NKKNLN10}, also referred to as estimation with partial knowledge \citep{ME11}, with many/few labeled examples (Section~\ref{sec:SharedRep}). (e),(f)~Multi-domain training with many/few labeled examples from the unobserved domain (Section~\ref{sec:RegSideInfo}).}
\label{fig:XYZsingle}
\end{figure*}

In terms of testing, we treat two tasks. The first is \emph{multi-domain regression}, in which the algorithm is asked to predict $\by$ based on an observation of $\bx_1$ and $\bx_2$. The second is \emph{single-domain regression}, where prediction should be based solely on $\bx_1$ (including the case where no $\bx_1$ labeled data is available for training, that is, $L_1=0$). Several archetypical situations are depicted in figs.~\ref{fig:XYZmulti} and~\ref{fig:XYZsingle}. Here, single- and double-lined circles correspond, respectively, to RVs that are unobserved and observed during testing. A continuous line, a dashed line, and lack of a line between circles corresponds, respectively, to many, few and zero training examples.

\section{Estimation Theoretic Formulation}
\label{sec:EstTheoForm}
In this paper we adopt and generalize the framework proposed in \cite{ME11} by posing our problem as one of estimation with partial knowledge of statistical relations. Before formalizing our multi-domain semi-supervised problem in estimation theoretic terms, we first recall the common practice for regression from one domain with a limited number of examples.

\subsection{Single-Domain Regression}
\label{sec:SingleDomRegSingleDomTrain}
Suppose we are given a sample $\{\bx^\ell,\by^\ell\}_{\ell=1}^L$, $\bx \in \RN^M$, independently drawn from the joint distribution of $X$ and $Y$. If $L$ is very large, then nonparametric methods can be used to approximate the conditional expectation estimator $\varphi(\bx)=\EE[Y|X=\bx]$ with great accuracy at any $\bx$. Such estimates, however, are often far from accurate when $L$ is small. Common practice in such situations is to use parametric or semi-parametric methods that impose some structure on the sought predictor. In other words, rather than trying to approximate the regression function $\varphi(\bx)=\EE[Y|X=\bx]$, which minimizes the mean square error among all functions of $X$, we settle for approximating the optimal predictor among some family $\A$ of functions:
\begin{equation}\label{eq:A_Optimal_Est}
\varphi_{\A}=\argmin_{\varphi\in\A}\EE\left[\|Y-\varphi(X)\|^2\right].
\end{equation}
The less rich the class $\A$ is, the more accurate we can typically approximate $\varphi_{\A}(X)$ from the training data. This comes, of course, at the cost that the (theoretical) MSE that $\varphi_{\A}(X)$ achieves is higher. This is the well known bias-variance tradeoff. In the sequel, we term the function $\varphi_{\A}(X)$ of \eqref{eq:A_Optimal_Est} the $\A$-optimal estimator of $Y$ from $X$.

One of the simplest structural restrictions corresponds to linear estimation, so that $\A$ is the set of all linear functions from $\RN^M$ to $\RN^N$. In this case,
\begin{equation}
\varphi_{\A}(X)=\bGa_{YX}\bGa_{XX}^\dag X.
\end{equation}
The second-order moment matrices $\bGa_{YX},\bGa_{XX}$ can be estimated from the training set, for example, by using sample moments. A more general model corresponds to functions of the form
\begin{equation}
\varphi(X)=\sum_{k=1}^K a_k \varphi_k(X),
\end{equation}
where $\{\varphi_k\}_{k=1}^K$ is a predefined set of functions and the coefficients $\{a_k\}_{k=1}^K$ are arbitrary. The optimal set of coefficients $\ba=\begin{pmatrix}a_1 & \cdots a_K\end{pmatrix}^T$ is given in this case by
\begin{equation}\label{eq:aParametricSingleView}
\ba = \bGa_{\Phi\Phi}^\dag \bGa_{\Phi Y},
\end{equation}
where $\bGa_{\Phi\Phi}$ denotes the $K\times K$ matrix whose $(i,j)$-th entry is $\EE[\varphi_i^T(X)\varphi_j(X)]$ and $\bGa_{\Phi Y}$ is a $K\times 1$ vector whose $i$th component is $\EE[\varphi_i^T(X)Y]$. These quantities can be estimated from the training data similar to the linear setting.

In both examples above, $\A$ forms a linear subspace of functions: for every $\varphi^1,\varphi^2\in\A$ and $\alpha,\beta\in\RN$, the function $\alpha\varphi^1+\beta\varphi^2$ also belongs to $\A$. For future reference, we note that this claim is also trivially true when $\A$ is taken to be the set of all (Borel-measurable) functions, in which case $\varphi_{\A}(X)=\EE[Y|X]$, and when $\A$ contains only the zero function, in which case $\varphi_{\A}(X)=0$.

\subsection{Statistical Knowledge Deduced from Separate Training Sets}
In our setting we have access to two sperate unpaired sets of labeled examples, one for each domain. Consequently, besides the standard uncertainty in statistics, which has to do with the fact that the underlying distributions are not known but rather only samples are observed, here there is another degree of uncertainty. Specifically, even if the number of training examples is taken to infinity in all three sets, we can only hope to be able to determine the joint distributions $F_{X_1Y}$, $F_{X_2Y}$ and $F_{X_1X_2}$. These do not suffice in general for computing the MMSE estimate $\EE[Y|X_1,X_2]$. To focus only on the second type of uncertainty, we assume that we are able to perform single domain regression from each of the training sets with very small variance (at the expense of possible bias). Specifically, we assume that we can determine the $\A$-optimal predictor of $Y$ given $X_1$ as well as the $\B$-optimal predictor of $Y$ from $X_2$, where $\A$ and $\B$ are classes of functions chosen in accordance with the cardinality of the two sets. Note that each of the single-domain predictors may be very poor. In particular, if there are no labeled training examples from one of the domains then we choose the corresponding class of valid predictors to contain only the zero function. Therefore, if, for instance, we have $L_1=0$ labeled examples from domain $X_1$, then we set $\A=\{0\}$ so that the $\A$-optimal predictor of $Y$ given $X_1$ is simply $\varphi_{\A}(X_1)=0$.

\begin{figure}[t]\centering
\includegraphics[scale=1]{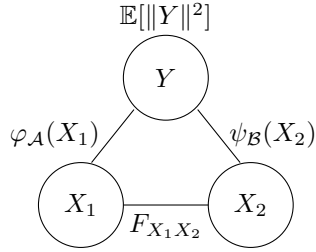}
\caption{Known statistical relationships. Each of the single-domain predictors may perform arbitrarily poorly (in particular, it is possible that $\varphi_{\A}(X_1)=0$ or $\psi_{\B}(X_2)=0$).}
\label{fig:SemiHiddenRelStatKnow}
\end{figure}

We further assume that the existence of many unlabeled examples $(X_1,X_2)$ allows accurately determining the joint distribution of $X_1$ and $X_2$, for example, using nonparametric methods. Finally, we assume that there are enough labeled examples from at least one of the domains such that the second-order moment of $Y$ can be accurately estimated. The statistical relationships assumed known are depicted in Fig.~\ref{fig:SemiHiddenRelStatKnow}.

In a more mathematical language, assume we are given two functions $\varphi_{\A}:\RN^{M_1}\rightarrow\RN^N$ and $\psi_{\B}:\RN^{M_2}\rightarrow\RN^N$, a cumulative probability function $F_{X_1X_2}$ over $\RN^{M_1\times M_2}$ and a scalar $c>0$. Then, what we know regarding the RVs $X_1$, $X_2$ and $Y$ is that their distribution $F_{X_1X_2Y}$ belongs to the set $\F$ of distributions satisfying
\begin{align}\label{eq:setF}
&\varphi_{\A}=\argmin_{\varphi\in\A}\EE[\|Y-\phi(X_1)\|^2], \quad
\psi_{\B}=\argmin_{\psi\in\B}\EE[\|Y-\psi(X_2)\|^2], \nonumber\\
&F_{X_1X_2Y}(\bx_1,\bx_2,\infty)=F_{X_1X_2}(\bx_1,\bx_2), \quad \EE[\|Y\|^2]=c.
\end{align}
We assume throughout the paper that $\A$ and $\B$ form linear subspaces of functions, as discussed in Section~\ref{sec:SingleDomRegSingleDomTrain}.

As an illustrative example, suppose that $X_1$, $X_2$ and $Y$ are scalar RVs, and that $\A$ and $\B$ are the sets of all linear functions from $\RN$ to $\RN$. Assume further that we know that the best linear estimator of $Y$ from $X_1$ is $\varphi_{\A}(X_1) = 0.1X_1$, the best linear estimator of $Y$ from $X_2$ is $\psi_{\B}(X_2)=0.2X_2$, the probability density function (pdf) of $(X_1,X_2)$ is $f_{X_1X_2}(x_1,x_2)\propto\exp\{-(x_1^2+x_2^2)/2\}$, and that $\EE[Y^2]=1$. Then the normal density
\begin{equation}
f_{X_1X_2Y}(x_1,x_2,y)\propto\exp\left\{-\frac{1}{2}\begin{pmatrix}x_1&x_2&y\end{pmatrix}\begin{pmatrix}
1   & 0   & 0.1\\
0   & 1   & 0.2\\
0.1 & 0.2 & 1
\end{pmatrix}^{-1}\begin{pmatrix}x_1\\x_2\\y\end{pmatrix}\right\}
\end{equation}
qualifies with all these restrictions and is thus valid. In fact, there is an infinite number (a continuum) of other feasible densities. For instance, it can be easily verified that the Gaussian mixture pdf
\begin{align}
f_{X_1X_2Y}(x_1,x_2,y)&\propto\exp\left\{-\frac{1}{2}\begin{pmatrix}x_1&x_2&y\end{pmatrix}\begin{pmatrix}
1   & 0   & 0.2\\
0   & 1   & 0\\
0.2 & 0   & 1
\end{pmatrix}^{-1}\begin{pmatrix}x_1\\x_2\\y\end{pmatrix}\right\} \nonumber\\
&\hspace{3cm}+\exp\left\{-\frac{1}{2}\begin{pmatrix}x_1&x_2&y\end{pmatrix}\begin{pmatrix}
1   & 0   & 0\\
0   & 1   & 0.4\\
0   & 0.4 & 1
\end{pmatrix}^{-1}\begin{pmatrix}x_1\\x_2\\y\end{pmatrix}\right\}
\end{align}
is also consistent with all the restrictions, making it a valid candidate as well. By contrast, the density
\begin{equation}
f_{X_1X_2Y}(x_1,x_2,y)\propto\exp\left\{-\frac{1}{2}\begin{pmatrix}x_1&x_2&y\end{pmatrix}\begin{pmatrix}
2   & 0   & 0.2\\
0   & 1   & 0.2\\
0.2 & 0.2 & 1
\end{pmatrix}^{-1}\begin{pmatrix}x_1\\x_2\\y\end{pmatrix}\right\}
\end{equation}
satisfies all requirements except for the demand that it be consistent with the given marginal distribution $f_{X_1X_2}(x_1,x_2)$. Therefore, it is not feasible.

\subsection{Goals}
The first problem we address in this paper is multi-domain regression. In this context, we would like to construct a predictor of $Y$ from the two domains $X_1$ and $X_2$, where the only knowledge we have is that $F_{X_1X_2Y}\in\F$. The second problem we tackle is single-domain regression. Here, the goal is to construct an estimator of $Y$ given $X_1$ alone based, again, only on the knowledge that $F_{X_1X_2Y}\in\F$. The special case of \emph{shared-representation learning}, in which no labeled examples from the first domain are available, corresponds to setting $\A=\{0\}$. The setting of \emph{cross modality learning}, in which there is no access to training examples from the second domain, can be addressed by setting $\B=\{0\}$. The general case we treat here can account for a wide spectrum of possibilities, including these two extremes.

Any predictor of $Y$, whether a function of $X_1$ and $X_2$ or of $X_1$ alone, may perform well under certain distributions $F_{X_1X_2Y}\in\F$ and worse under others. Our goal is therefore to uniformly minimize the MSE over $\F$. As we will see, this minimax approach leads to simple closed form solutions, which can be easily applied to the various settings discussed in Section~\ref{sec:ProbForm}.

\section{Multi-Domain Regression}
\label{sec:MultiDomainRegression}

Assume that the joint distribution of the triplet $(X_1,X_2,Y)$ is known to belong to the family $\F$ of \eqref{eq:setF}, where $\A$ and $\B$ are linear subspaces of prediction functions. For any distribution $F_{X_1X_2Y}$, the MSE attained by an estimator $\hat{Y}=\rho(X_1,X_2)$ is defined as
\begin{equation}
{\rm MSE}(F_{X_1X_2Y},\rho) = \EE\left[\|Y-\rho(X_1,X_2)\|^2\right],
\end{equation}
where the expectation is with respect to $F_{X_1X_2Y}$. Since the MSE depends on $F_{X_1X_2Y}$, which is unknown, our approach is to seek the estimator whose worst-case MSE over $\F$ is minimal. This minimax concept is widely practiced in deterministic parameter estimation \citep{EBN04,E08} as well as in random parameter estimation \citep{EM04,EM05}. More concretely, we are interested in\footnote{The subscript `M' stands for `multi-domain.'}
\begin{equation}\label{eq:RhoStarMinimaxYZ}
\rho_{\rm M}=\argmin_{\rho}\sup_{F_{X_1X_2Y}\in\F}{\rm MSE}(F_{X_1X_2Y},\rho).
\end{equation}
The next theorem, whose proof can be found in Appendix~\ref{sec:ProofThmMxMultiview}, provides a means for solving this problem.

%

\begin{theorem}[Multi-domain minimax-MSE prediction]\label{thm:MxMultiview}
Choose any distribution $F_{X_1X_2Y}\in\F$ and consider the estimator
\begin{equation}\label{eq:RhoStarMinimumYZ}
\rho_{\C} = \argmin_{\rho\in\C}{\rm MSE}(F_{X_1X_2Y},\rho),
\end{equation}
where $\C=\A+\B$, namely
\begin{equation}
\C=\left\{\rho\,:\,\rho(\bx_1,\bx_2)=\phi(\bx_1)+\psi(\bx_2),\, \phi\in\A,\, \psi\in\B\right\}.
\end{equation}
Then
\begin{enumerate}
\item \label{ItemRhoStarInvariant}the function $\rho_{\C}$ does not depend on the choice of $F_{X_1X_2Y}\in\F$;
\item \label{ItemMSEInvariant}the value ${\rm MSE}(F_{X_1X_2Y},\rho_{\C})$ does not depend on the choice of $F_{X_1X_2Y}\in\F$;
\item \label{ItemRhoStarMinimax}the estimator $\rho_{\C}$ of \eqref{eq:RhoStarMinimumYZ} is also the solution $\rho_{\rm M}$ to \eqref{eq:RhoStarMinimaxYZ}.
\end{enumerate}
\end{theorem}

Theorem~\ref{thm:MxMultiview} shows that instead of solving the minimax problem \eqref{eq:RhoStarMinimumYZ}, we can equivalently solve the minimization problem \eqref{eq:RhoStarMinimaxYZ}. Namely, all we need to do is determine the MMSE estimator of $Y$ among all functions of the form $\phi(X_1)+\psi(X_2)$ with $\phi\in\A$ and $\psi\in\B$. The importance of this observation follows from the fact that, as we show below, for many practical cases, the latter possesses a simple closed form solution.

Before demonstrating the utility of the minimax MSE approach, we note that optimizing the worst-case performance of an estimator is very conservative and may sometimes lead to over-pessimistic solutions. As an alternative, researchers in many application areas have proposed minimizing the worst-case \emph{regret} \citep{EBN04,EM04,ME10,ME11}. The regret of an estimator $\rho(X_1,X_2)$ is defined as the difference between the MSE it achieves and the MSE of the MMSE solution, namely
\begin{align}
{\rm REG}(F_{X_1X_2Y}, \rho) &= \EE\left[\|Y-\rho(X_1,X_2)\|^2\right] - \EE\left[\|Y-\EE[Y|X_1,X_2]\|^2\right].
\end{align}
In this expression, both terms depend on $F_{X_1X_2Y}$, so that minimization of the worst-case regret is generally not equivalent to minimization of the worst-case MSE. Additional insight into the regret can be obtained from its equivalent characterization \citep{ME11} as the MSE between $\rho(X_1,X_2)$ and $\EE[Y|X_1,X_2]$, namely
\begin{align}\label{eq:RegretDefinition2}
{\rm REG}(F_{X_1X_2Y},\rho) &= \EE\left[\|\rho(X_1,X_2)-\EE[Y|X_1,X_2]\|^2\right].
\end{align}
As we show in the following theorem, however, in the multi-domain prediction setting, the minimax-regret estimator coincides with the minimax-MSE solution. The proof of the theorem is provided in Appendix~\ref{sec:ProofThmMxReg}.

\begin{theorem}[Multi-domain minimax-regret prediction]\label{thm:MxRegMultiview}
Consider the problem
\begin{equation}
\rho_{\rm R} = \argmin_{\rho}\sup_{F_{X_1X_2Y}\in\F}{\rm REG}(F_{X_1X_2Y},\rho),
\end{equation}
where minimization is performed over all functions $\rho$ of $X_1$ and $X_2$. Then its solution $\rho_{\rm R}$ coincides with $\rho_{\rm M}$ of \eqref{eq:RhoStarMinimaxYZ}.
\end{theorem}

We now apply Theorem~\ref{thm:MxMultiview} in several scenarios.

\subsection{Single-Domain Training}
\label{sec:SingleDomTrainMultiDomReg}
Consider the situation of figs.~\ref{fig:XYZmulti}\subref{fig:MultiDomReg_SingleDomTrainMany} and~\ref{fig:XYZmulti}\subref{fig:MultiDomReg_SingleDomTrainFew}, where we have at our disposal only labeled examples from one domain, say $X_1$. In this case $\B=\{0\}$ so that $\C=\A$. Consequently, the solution to \eqref{eq:RhoStarMinimumYZ} is simply
\begin{equation}
\rho_{\C}(X_1,X_2)=\varphi_{\A}(X_1).
\end{equation}
This shows that in coming to label unseen examples, there is no gain in basing the prediction on the domain $X_2$ for which we have no labeled training examples. Furthermore, at least from a worst-case perspective, there is no better strategy than using our initial predictor based on $X_1$ alone. More concretely, for any estimator that differs from $\varphi_{\A}(X_1)$ (and in particular one that is a function of $X_2$), there exist distributions $F_{X_1X_2Y}\in\F$ (one maybe being the true underlying distribution) under which the predictor $\varphi_{\A}(X_1)$ performs better.

This result does not stand in contrast to the basic observation in multi-view learning that unlabeled data helps \citep{BM98}. This is because in our setting, we do not assume that the two views are ``coherent'' or tend to agree in any sense, as done, for instance, in \cite{KF07} in the context of multi-view regression.

\subsection{Multi-Domain Linear Regression}
\label{sec:MultiDomLinReg}
Suppose, as in Fig.~\ref{fig:XYZmulti}\subref{fig:MultiDomReg_MultiDomTrainParam}, that we have a limited amount of labeled examples from both domains, which only suffice for identifying (with very high precision) the optimal linear predictor from each view. In this case $\A$ and $\B$ correspond to the collection of all linear functions from $\RN^{M_1}$ to $\RN^N$ and from $\RN^{M_2}$ to $\RN^N$, respectively. Consequently, $\C$ is the set of all linear functions from $\RN^{M_1}\times\RN^{M_2}$ to $\RN^N$. This implies that the solution to \eqref{eq:RhoStarMinimumYZ} is simply the best linear predictor of $Y$ based on $X_1$ and $X_2$, namely
\begin{equation}\label{eq:MultiDomainLinear}
\rho_{\C}(X_1,X_2)=
\begin{pmatrix}
\bGa_{YX_1} & \bGa_{YX_2}
\end{pmatrix}
\begin{pmatrix}
\bGa_{X_1\!X_1} & \bGa_{X_1\!X_2} \\
\bGa_{X_2\!X_1} & \bGa_{X_2\!X_2}
\end{pmatrix}^\dag
\begin{pmatrix}
X_1 \\ X_2
\end{pmatrix}.
\end{equation}
The second-order moments $\bGa_{X_i\!X_j}$, $i,j\in\{1,2\}$, can be estimated from the unlabeled training set. Similarly, the matrices $\bGa_{Y\!X_j}$, $i,j\in\{1,2\}$, can be determined from the labeled sets.

The dependence of the multi-domain predictor $\rho_{\C}$ on the single-domain estimators $\phi_{\A}$ and $\psi_{\B}$ is not apparent at first sight. However, recall that the orthogonality principle states that $\EE[(Y-\phi_{\A}(X_1))X_1^T]=0$ and $\EE[(Y-\psi_{\B}(X_2))X_2^T]=0$. Therefore, the terms $\bGa_{YX_1}$ and $\bGa_{YX_2}$ in \eqref{eq:MultiDomainLinear} can be replaced by $\EE[\phi_{\A}(X_1)X_1^T]$ and $\EE[\psi_{\B}(X_2)X_2^T]$, respectively. As these expectations are with respect to $F_{X_1}$ and $F_{X_2}$, their computation can be carried out based only on the knowledge of $F_{X_1X_2}$, $\phi_{\A}$ and $\psi_{\B}$, which is available according to our problem formulation.

\subsection{Multi-Domain Parametric Regression}
\label{sec:MultiDomParReg}
The above observation naturally extends to the case in which the training sets suffice for identifying the optimal parametric predictors of the forms
\begin{equation}
\varphi(X_1)=\sum_{k=1}^{K_1} a^1_k \varphi_k(X_1),\quad \psi(X_2)=\sum_{k=1}^{K_2} a^2_k \psi_k(X_2),
\end{equation}
where $\{\varphi_k\}_{k=1}^{K_1}$ and $\{\psi_k\}_{k=1}^{K_2}$ are given functions and $\{a^1_k\}_{k=1}^{K_1}$ and $\{a^2_k\}_{k=1}^{K_2}$ are arbitrary parameters. In this situation, $\C$ corresponds to the family of functions having the form
\begin{equation}
\rho(X_1,X_2)=\sum_{k=1}^{K_1} a^1_k \varphi_k(X_1)+\sum_{k=1}^{K_2} a^2_k \psi_k(X_2).
\end{equation}
Thus, the optimal set of parameters $\ba=\begin{pmatrix}a^1_1 & \cdots & a^1_{K_1} & a^2_1 & \cdots & a^2_{K_2}\end{pmatrix}^T$ is given by
\begin{equation}
\ba^*=
\begin{pmatrix}
\bGa_{\Phi\Phi} & \bGa_{\Phi\Psi} \\
\bGa_{\Psi\Phi} & \bGa_{\Psi\Psi} \\
\end{pmatrix}^\dag
\begin{pmatrix}
\bGa_{\Phi Y} \\ \bGa_{\Psi Y}
\end{pmatrix},
\end{equation}
with $\bGa_{\Phi\Phi}$, $\bGa_{\Psi\Psi}$, $\bGa_{\Phi Y}$ and $\bGa_{\Psi Y}$ being as in \eqref{eq:aParametricSingleView} and $\bGa_{\Phi\Psi}$ being a $K_1\times K_2$ matrix whose $(i,j)$-th entry is $\EE[\varphi_i(Y)^T\psi_j(Z)]$. Similar to linear regression, the vectors $\bGa_{\Phi Y}$ and $\bGa_{\Psi Y}$ can be replaced, due to the orthogonality principle, by vectors whose $j$-th entries are $\EE[\varphi_j^T(X_1)\varphi_{\A}(X_1)]$ and $\EE[\psi_j^T(X_1)\psi_{\B}(X_2)]$, respectively.


\subsection{Multi-Domain Partially Linear Regression}
\label{sec:MultiDomPartialLinReg}
Suppose, as in Fig.~\ref{fig:XYZmulti}\subref{fig:MultiDomReg_MultiDomTrainSemiParam}, that we have numerous labeled examples from the first domain, allowing us to determine $\EE[Y|X_1]$, and only a limited amount of examples from the second domain, so that we can only determine the best linear predictor of $Y$ from $X_2$. In this setting, Theorem~\ref{thm:MxMultiview} implies that the minimax-optimal predictor based on $X_1$ and $X_2$ is the estimator minimizing the MSE among all functions of the form
\begin{equation}
\rho(X_1,X_2) = \ba(X_1) + \bB X_2,
\end{equation}
where $\ba:\RN^{M_1}\rightarrow\RN^N$ is an arbitrary function and $\bB\in\RN^{N\times M_2}$ is some matrix. It was shown in \cite{MSE11} that the solution to this particular case is given by
\begin{equation}\label{eq:PLMMSE}
\rho_{\rm M}(X_1,X_2) = \EE[Y|X_1] + \bGa_{YW}\bGa_{WW}^\dag W,
\end{equation}
where $W=X_2-\EE[X_2|X_1]$.

The intuition here is that we need to make sure we do not account for variations in $Y$ twice when fusing information from $X_1$ and $X_2$. Thus, we start with the estimate $\varphi_{\A}(X_1)=\EE[Y|X_1]$, and then update it with the LMMSE estimate of $Y$ based on the \emph{innovation} $X_2-\EE[X_2|X_1]$ of $X_2$ with respect to $\varphi_{\A}(X_1)$.

In practice, the term $\EE[Y|X_1]$ can be approximated from the labeled training examples of the first domain, \eg using nonparametric methods. The second term in \eqref{eq:PLMMSE} can be obtained via a three-stage procedure. Specifically, we first employ a nonparametric technique to approximate $\xi(\bx_1)=\EE[X_2|X_1=\bx_1]$ from the unlabeled set. Next, we use the unlabeled samples to form the set $\{\xi(\bx_1^u),\bx_2^u\}_{u=L_1+L_2+1}^{L_1+L_2+U}$, from which we approximate the covariance matrix $\bGa_{WW}$ of $W=X_2-\EE[X_2|X_1]$. Lastly, we approximate $\bGa_{YX_2}$ from the labeled examples $\{\bx_2^\ell,\by^\ell\}_{\ell=L_1+1}^{L_1+L_2}$ and $\bGa_{Y\xi(X_1)}$ from the labeled examples $\{\xi(\bx_1^\ell),\by^\ell\}_{\ell=1}^{L_1}$ in order to compute $\bGa_{YW}=\bGa_{YX_2}-\bGa_{Y\xi(X_1)}$.

\subsection{Multi-Domain Semi-Parametric Regression}
\label{sec:MultiDomSemiParReg}
Suppose as above, that we know $\EE[Y|X_1]$, however we can also determine the best estimator of $Y$ from $X_2$ among the parametric family
\begin{equation}\label{eq:psiParameteric}
\psi(X_2)=\sum_{k=1}^{K} a_k \psi_k(X_2).
\end{equation}
In this case, according to Theorem~\ref{thm:MxMultiview}, the minimax-optimal estimator of $Y$ based on $X_1$ and $X_2$ is the one minimizing the MSE among all functions of the form
\begin{equation}\label{eq:semiParameteric}
\rho(X_1,X_2) = \ba(X_1) + \sum_{k=1}^{K} a_k \psi_k(X_2).
\end{equation}
The solution to this problem can be deduced by relying on the concept of $(\A,\B)$-innovation, as we now define.

\begin{definition}\label{def:ABinnovation}
The $(\A,\B)$ innovation of $X_2$ with respect to $X_1$, which we denote by $\rho_{\A,\B}(X_1,X_2)$, is the MMSE estimator of $Y$ among all functions of the form
\begin{equation}\label{eq:ABinnovation}
\psi(X_2)-\eta_{\psi}(X_1),
\end{equation}
with $\psi$ being some function in $\B$ and $\eta_{\psi}(X_1)$ denoting the $\A$-optimal estimator of $\psi(X_2)$ from $X_1$.
\end{definition}
Using this definition, we make the following observation regarding the structure of the minimax estimator, the proof of which is given in Appendix~\ref{sec:ProofSemiParametric}.
\begin{theorem}\label{thm:SemiParametric}
The solution to problem \eqref{eq:RhoStarMinimumYZ} can be expressed as
\begin{equation}\label{eq:SemiParametric}
\rho_{\C}(X_1,X_2) = \varphi_{\A}(X_1) + \rho_{\A,\B}(X_1,X_2),
\end{equation}
where $\rho_{\A,\B}(X_1,X_2)$ is the $(\A,\B)$-innovation of $X_2$ with respect to $X_1$.
\end{theorem}

In our setting, $\A$ corresponds to the set of all functions from $\RN^{M_1}$ to $\RN^N$ so that $\varphi_{\A}(X_1)=\EE[Y|X_1]$. Furthermore, $\B$ is the family of functions from $\RN^{M_2}$ to $\RN^N$ having the form \eqref{eq:psiParameteric}. Therefore, for any $\psi\in\B$, the $\A$-optimal estimator of $\psi(X_2)$ based on $X_1$ is given by
\begin{align}
\eta_{\psi}(X_1) = \EE[\psi(X_2)|X_1]
= \EE\left[\left.\sum_{k=1}^{K} a_k \psi_k(X_2)\right|X_1\right]
= \sum_{k=1}^{K} a_k \EE[\psi_k(X_2)|X_1].
\end{align}
Consequently, $\rho_{\A,\B}(X_1,X_2)$ in \eqref{eq:SemiParametric} is of the form
\begin{align}
\psi(X_2)-\eta_{\psi}(X_1) = \sum_{k=1}^{K} a_k \psi_k(X_2) - \sum_{k=1}^{K} a_k \EE[\psi_k(X_2)|X_1]
= \sum_{k=1}^{K} a_k \rho_k(X_1,X_2),
\end{align}
where we denoted $\rho_k(X_1,X_2)=\psi_k(X_2)-\EE[\psi_k(X_2)|X_1]$. The optimal set of coefficients is given by
\begin{equation}\label{eq:a_rho_x}
\ba^*= \bGa_{\rho\rho} \bGa_{\rho Y}
\end{equation}
where $\bGa_{\rho\rho}$ and $\bGa_{\rho Y}$ are as in \eqref{eq:aParametricSingleView} with $\varphi_i(X_1)$ replaced by $\rho_i(X_1,X_2)$.

To conclude, the optimal estimator of the form \eqref{eq:semiParameteric} is
\begin{equation}
\rho_{\rm M}(X_1,X_2) = \EE[Y|X_1] + \sum_{k=1}^{K} a_k \left(\psi_k(X_2) - \EE[\psi_k(X_2)|X_1]\right),
\end{equation}
with coefficients $\{a_k\}$ given by \eqref{eq:a_rho_x}. The first term in this expression can be approximated via nonparametric regression techniques from the labeled training examples of the first domain. The second term can be computed in two stages. First, each of the functions $\{\psi_k(X_2)\}_{k=1}^K$ is regressed on $X_1$ using the unlabeled data set, to obtain an approximation of $\EE[\psi_k(X_2)|X_1]$. Then, $Y$ is linearly regressed against $\{\psi_k(X_2) - \EE[\psi_k(X_2)|X_1]\}_{k=1}^K$, using the two labeled sets, as discussed in Section~\ref{sec:MultiDomPartialLinReg}.

\section{Single-Domain Regression with Multi-Domain Training}
\label{sec:SingleDomainRegression}
Next, we address the setting in which at the testing stage our predictor is only supplied with one type of features, say $X_1$. The interesting question in this context is how to take into account the training sets of both domains in order to design an improved estimator of $Y$ based on $X_1$ alone.

Since our estimator operates on $X_1$ and is judged by the proximity of its output to $Y$, its performance is only affected by the joint distribution of $Y$ and $X_1$. It may thus seem at first that the second set of features $X_2$ cannot be of help in improving estimation accuracy. However, note that $F_{X_1Y}$ is not fully known in our setting. Thus, being told the statistical relations between $Y$ and $X_2$ and between $X_1$ and $X_2$, might help to narrow down the set of candidate distributions $F_{X_1Y}$ for which we need to design an estimator.

The statistical relations known to us are the same as in Section~\ref{sec:MultiDomainRegression}. Namely, we know that $F_{X_1X_2Y}$ belongs to the class $\F$ of \eqref{eq:setF}. Therefore, as in Section~\ref{sec:MultiDomainRegression}, our goal is to optimize the worst case performance of our estimator over $\F$. As it turns out, in contrast with the multi-domain problem, in the single-domain setting the minimax MSE and minimax regret solutions no longer coincide. Here, we focus on minimizing the worst-case regret. As will be clear from the proof provided in Appendix \ref{thm:MxRegSingleview}, determining the minimax-MSE estimator in the single-domain setting is much harder than minimizing the worst-case regret. The former remains an open problem.

In single domain regression, whatever we do, our estimator will not achieve lower MSE than the conditional expectation $\EE[Y|X_1]$. Therefore, the \emph{regret} of interest is now
\begin{align}
{\rm REG}(F_{X_1X_2Y},\rho) &= \EE\left[\|Y-\rho(X_1)\|^2\right] - \EE\left[\|Y-\EE[Y|X_1]\|^2\right].
\end{align}
As in the multi-domain setting, this regret here can be written as \citep{ME11}
\begin{align}
{\rm REG}(F_{X_1X_2Y},\rho) &= \EE\left[\|\rho(X_1)-\EE[Y|X_1]\|^2\right].
\end{align}
Our goal is to determine the minimax-regret estimator\footnote{The subscript `S' stands for `single-domain.'}
\begin{equation}\label{eq:RhoStarMinimaxY}
\rho_{\rm S}=\argmin_{\rho}\sup_{F_{X_1X_2Y}\in\F}{\rm REG}(F_{X_1X_2Y},\rho),
\end{equation}
where now minimization is performed only over functions $\rho$ of $X_1$.

The next theorem, whose proof may be found in Appendix~\ref{sec:ProofThmMxReg}, describes the single-domain minimax-regret estimator in terms of the multi-domain minimax-MSE solution.
\begin{theorem}[Single-domain minimax-regret prediction]\label{thm:MxRegSingleview}
The solution to problem \eqref{eq:RhoStarMinimaxY} is given by
\begin{equation}
\rho_{\rm S}(X_1) = \EE[\rho_{\rm M}(X_1,X_2)|X_1],
\end{equation}
where $\rho_{\rm M}(X_1,X_2)$ is the multi-domain minimax estimator \eqref{eq:RhoStarMinimaxYZ}.
\end{theorem}

This result has a very simple and intuitive explanation. We know that $F_{X_1X_2Y}$ belongs to the set $\F$, and therefore $\rho_{\rm M}(X_1,X_2)$ is the optimal estimate of $Y$ in a minimax-MSE sense. However, we cannot use this estimate as it is a function of $X_2$, which is not measured in our setting. What Theorem~\ref{thm:MxRegSingleview} shows is that the optimal strategy is to estimate $\rho_{\rm M}(X_1,X_2)$ based on the available measurements, which are $X_1$ alone. Computation of the conditional expectation $\EE[\rho_{\rm M}(X_1,X_2)|X_1]$ only requires knowledge of the marginal distribution $F_{X_1X_2}$, which is available in our setting.

We now apply this result to two interesting special cases.

\subsection{Cross Domain Regression}
\label{sec:CrossDomain}
In cross-modality learning \citep{NKKNLN10}, we only have labeled examples from domain $X_1$ and not from $X_2$, as illustrated in figs.~\ref{fig:XYZsingle}\subref{fig:SingleDomReg_CrossModMany} and~\ref{fig:XYZsingle}\subref{fig:SingleDomReg_CrossModFew}. The basic intuition here, as presented in \cite{NKKNLN10}, is that the unlabeled data may be used to boost the performance of the best single-domain estimator $\varphi_{\A}(X_1)$ that can be designed based solely on labeled examples from the domain $X_1$.

This setting can be treated within our framework by setting $\psi_{\B}(X_2)=0$. As we have seen in Section~\ref{sec:SingleDomTrainMultiDomReg}, in this situation $\rho_{\rm M}(X_1,X_2)=\varphi_{\A}(X_1)$. Therefore, the single-domain minimax-regret predictor of $Y$ from $X_1$ is given by
\begin{equation}
\rho_{\rm S}(X_1) = \EE[\varphi_{\A}(X_1)|X_1] = \varphi_{\A}(X_1).
\end{equation}
We see that despite the fact that we know $F_{X_1X_2}$, there is no better strategy than using the estimator $\varphi_{\A}(X_1)$ here. This implies that cross-modality learning is not useful unless additional knowledge on the underlying distributions is available.

The authors of \cite{NKKNLN10} used cross-modality learning to classify isolated words from either audio or video (lipreading). It was reported that unlabeled audio-visual examples helped improve visual recognition but failed to boost the performance of an audio classifier. This empirical result aligns with our theoretical analysis, which states that, in the worst-case scenario, there is nothing better to do than disregarding the modality for which no labeled examples are available.

\subsection{Shared Representation Regression}
\label{sec:SharedRep}
In shared-representation learning \citep{NKKNLN10}, also referred to as estimation with partial knowledge \citep{ME11}, we have no labeled examples from domain $X_1$ but rather only from $X_2$. This is illustrated in figs.~\ref{fig:XYZsingle}\subref{fig:SingleDomReg_SharedRepMany} and~\ref{fig:XYZsingle}\subref{fig:SingleDomReg_SharedRepFew}. Since we can learn a predictor $\psi_{\B}(X_2)$ from the second domain, and only measure an instance $X_1$ from the first domain, a naive approach would be to feed the predictor $\psi_{\B}$ with an estimate of $X_2$, which is based on $X_1$, rather than with $X_2$ itself. For example, the MMSE estimate $\EE[X_2|X_1]$ can be approximated by nonparametric methods from the unlabeled training set. However, as we now show, this strategy is generally \emph{not} minimax-optimal.

Recall from Section~\ref{sec:SingleDomTrainMultiDomReg} that the multi-domain predictor corresponding to the setting in which $\A=\{0\}$ is $\rho_{\rm M}(X_1,X_2)=\psi_{\B}(X_2)$. Therefore, the single-domain minimax-regret predictor of $Y$ from $X_1$ is given by
\begin{equation}\label{eq:sharedRepEst}
\rho_{\rm S}(X_1) = \EE[\psi_{\B}(X_2)|X_1]
\end{equation}
in this case. This solution generalizes the estimator of \cite[Thm.~8]{ME11}, which was developed for the case in which $\B$ is the set of all functions. In the latter scenario, $\psi_{\B}(X_2)=\EE[Y|X_2]$, and the two methods coincide.

As an example, consider the setting in which we have a limited number of labeled examples from domain $X_2$, which only allows to determine the best linear predictor of $Y$ from $X_2$. In this case, $\psi_{\B}(X_2)=\bGa_{YX_2}\bGa_{X_2X_2}^\dag X_2$, implying that $\rho_{\rm S}(X_1) = \EE[\bGa_{YX_2}\bGa_{X_2X_2}^\dag X_2|X_1]=\bGa_{YX_2}\bGa_{X_2X_2}^\dag \EE[X_2|X_1]$. Namely, minimax-regret estimation does boil down, in this setting, to the naive strategy of applying $\psi_{\B}$ on $\EE[X_2|X_1]$. This, however, is not always the case. Suppose, for instance, that we have numerous examples from domain $X_2$, so that $\B$ is the set of all functions from $\RN^{M_2}$ to $\RN^{N}$. In this situation, $\psi_{\B}(X_2)=\EE[Y|X_2]$, so that $\rho_{\rm S}(X_1) = \EE[\EE[Y|X_2]|X_1]$. This solution does not generally coincide with the naive estimator $\EE[Y|\EE[X_2|X_1]]$.

The estimator \eqref{eq:sharedRepEst} can be approximated from the available training data by first determining the function $\psi_{\B}(\bx_2)$ from the labeled set of the second domain and then using nonparametric regression on the set $\{\bx_1^u,\psi_{\B}(\bx_2^u)\}_{u=L_1+L_2+1}^{L_1+L_2+U}$.

\subsection{Regression with Side Information}
\label{sec:RegSideInfo}

The general setting in which we have training data from both domains can be treated by employing Theorem~\ref{thm:SemiParametric}. Specifically, when $\A$ and $\B$ are two arbitrary spaces of prediction functions, $\rho_{\rm M}(X_1,X_2)$ is given by \eqref{eq:SemiParametric}, and therefore
\begin{equation}\label{eq:GenCase}
\rho_{\rm S}(X_1) = \varphi_{\A}(X_1) + \EE[\rho_{\A,\B}(X_1,X_2)|X_1],
\end{equation}
where $\rho_{\A,\B}(X_1,X_2)$ is the $(\A,\B)$ innovation of $X_2$ with respect to $X_1$. This representation highlights the fact that the second labeled set and the unlabeled set come into play in the term $\EE[\rho_{\A,\B}(X_1,X_2)|X_1]$.

To understand when training data from an unobserved domain cannot help, we recall from Definition~\ref{def:ABinnovation} that $\rho_{\A,\B}(X_1,X_2)$ is of the form $\psi(X_2)-\eta_{\psi}(X_1)$, with $\psi\in\B$ and $\eta_{\psi}(X_1)$ being the $\A$-optimal estimate of $\psi(X_2)$ from $X_1$. Therefore, the second term in \eqref{eq:GenCase} vanishes if, for example,
\begin{equation}\label{eq:regInnovVanish}
\EE[\psi(X_2)|X_1]=\eta_{\psi}(X_1)
\end{equation}
for every $\psi\in\B$. Intuitively, this can happen if the class $\A$ of functions is very rich and/or the class $\B$ is not. As an example, if $\A$ is the set of all functions from $\RN^{M_1}$ to $\RN^N$ then $\eta_{\psi}(X_1)=\EE[\psi(X_2)|X_1]$, so that \eqref{eq:regInnovVanish} is satisfied, indicating that the training set from the second domain is not needed. Indeed, in this situation $\varphi_{\A}(X_1)=\EE[Y|X_1]$, meaning that we can already determine the MMSE predictor of $Y$ from $X_1$ using the first training set so that no potential improvement can be obtained using the second set.

As a more interesting example, suppose that the RVs $X_1$ and $X_2$ are jointly Gaussian, that $\B$ is the set of all linear functions from $\RN^{M_2}$ to $\RN^N$, and that $\A$ contains the set of all linear functions from $\RN^{M_1}$ to $\RN^N$. In this case, every $\psi\in\B$ corresponds to some matrix $\bA$ such that $\psi(X_2)=\bA X_2$. Consequently, using the fact that the MMSE estimate is linear in the Gaussian setting,
\begin{equation}\label{eq:GenCaseExamp1}
\EE[\psi(X_2)|X_1]=\EE[\bA X_2|X_1]=\bA \EE[X_2|X_1]=\bA\bGa_{X_2X_1}\bGa_{X_1X_1}^\dag X_1.
\end{equation}
Moreover, $X_1$ and $\psi(X_2)$ are jointly Gaussian, implying that
\begin{equation}\label{eq:GenCaseExamp2}
\eta_{\psi}(X_1) = \bGa_{\psi(X_2) X_1}\bGa_{X_1X_1}^\dag X_1 = \bA\bGa_{X_2X_1}\bGa_{X_1X_1}^\dag X_1.
\end{equation}
Thus, \eqref{eq:GenCaseExamp1} and \eqref{eq:GenCaseExamp2} coincide and \eqref{eq:regInnovVanish} is satisfied, indicating that the second training set is not required here as well.

Another interesting viewpoint can be obtained by switching the roles of $X_1$ and $X_2$ in the representation \eqref{eq:SemiParametric} of $\rho_{\rm M}(X_1,X_2)$. This leads to the expression
\begin{equation}\label{eq:GenCase2}
\rho_{\rm S}(X_1) = \EE[\psi_{\B}(X_2)|X_1] + \EE[\rho_{\B,\A}(X_2,X_1)|X_1].
\end{equation}
Here, we recognize the first term as being the shared-representation estimator \eqref{eq:sharedRepEst} of $Y$ from $X_1$, which does not use labeled examples from the domain $X_1$. Therefore, we see that the training set from the first (observed) domain is not needed if the second term in \eqref{eq:GenCase2} vanishes. Using the fact that $\rho_{\B,\A}(X_2,X_1)=\varphi(X_1)-\eta_{\varphi}(X_2)$ with $\varphi\in\A$ and $\eta_{\varphi}(X_2)$ being the $\B$-optimal estimate of $\varphi(X_1)$ from $X_2$, we conclude that this happens if, for example,
\begin{equation}\label{eq:regInnovVanish2}
\varphi(X_1)=\EE[\eta_{\varphi}(X_2)|X_1]
\end{equation}
for every $\varphi\in\A$. As a concrete example, consider again the setting in which the RVs $X_1$ and $X_2$ are jointly Gaussian and $\A$ and $\B$ are classes of linear functions. In this situation, $\varphi(X_1)=\bA X_1$ for some matrix $\bA$, so that $\eta_{\varphi}(X_2)=\bGa_{\varphi(X_1) X_2}\bGa_{X_2X_2}^\dag X_2=\bA\bGa_{X_1X_2}\bGa_{X_2X_2}^\dag X_2$ and, consequently,
\begin{equation}
\EE[\eta_{\varphi}(X_2)|X_1] = \bA\bGa_{X_1X_2}\bGa_{X_2X_2}^\dag\EE[X_2|X_1] = \bA\bGa_{X_1X_2}\bGa_{X_2X_2}^\dag\bGa_{X_2X_1}\bGa_{X_1X_1}^\dag X_1.
\end{equation}
Therefore, \eqref{eq:regInnovVanish2} is satisfied if $\bGa_{X_1X_2}\bGa_{X_2X_2}^\dag\bGa_{X_2X_1}\bGa_{X_1X_1}^\dag=\bI$, or, equivalently if $\bGa_{X_1X_1}-\bGa_{X_1X_2}\bGa_{X_2X_2}^\dag\bGa_{X_2X_1}=0$. The latter expression is no other than the error covariance of the MMSE estimate of $X_1$ from $X_2$. Therefore, condition \eqref{eq:regInnovVanish2} is satisfied in this setting if $X_1$ can be estimated from $X_2$ with no error. Indeed, in this scenario, we do not need to observe training examples from the domain $X_1$, as these can be synthetically generated from the examples of the second domain.

To approximate the resulting estimators from sets of points, it is often more convenient to use the form \eqref{eq:GenCase2} rather than \eqref{eq:GenCase}. As a concrete example, consider linear regression with nonlinear side information, namely where $\A$ is the set of all linear functions and $\B$ is the family of all (not necessarily linear) functions. Then, from Theorem~\ref{thm:MxRegSingleview} and \eqref{eq:PLMMSE} we conclude that
\begin{equation}\label{eq:linRegNonlinSideInfo}
\rho_{\rm S}(X_1) = \EE[\EE[Y|X_2]|X_1] + \bGa_{YW}\bGa_{WW}^\dag (X_1-\EE[\EE[X_1|X_2]|X_1]),
\end{equation}
where here $W=X_1-\EE[X_1|X_2]$. The terms $\EE[\EE[Y|X_2]|X_1]$ and $\EE[\EE[X_1|X_2]|X_1]$ can be approximated using nonparametric methods, similar to the discussion in Section~\ref{sec:SharedRep}, and the covariance matrices $\bGa_{YW}$ and $\bGa_{WW}$ can be approximated as in Section~\ref{sec:MultiDomPartialLinReg}.

\section{Experimental Results}
\label{sec:experiments}
We now demonstrate our regression approach, that derives from the theoretical results just presented, in two illustrative applications.

\subsection{Face Normalization}
Many facial recognition methods rely on a preprocessing stage, coined \emph{normalization}, which is aimed at removing variations that were not observed in the training database. These may include variations due to illumination, pose, facial expressions, and more. To demonstrate the utility of our approach, we now focus on the problem of producing a neutral expression face from a smiling one.

A straight forward way of tackling this problem is to learn a regression function from pairs of training images. This requires a database in which each subject appears at least twice, one time with a neutral expression and one time with a smile. Unfortunately, large data sets of this sort are hard to collect. In many practical situations one only has access to a database in which each subject appears only once. While different subjects may be wearing different expressions, direct inference of the statistical relation between a smiling and a neutral face is virtually impossible in such scenarios. To bypass this obstacle, we can use a second domain, or view, for which it is easy to obtain examples that are paired with the images in the database. This can be done, for example, by manually marking a set of points in several predefined locations on all images in the database. Thus, denoting by $(X_1,X_2,Y)$ a triplet of a smiling face, its point annotations, and the corresponding neutral expression image, we may construct an unlabeled set of annotated smiling faces $\{\bx_1^u,\bx_2^u\}$ and a set of annotated neutral expression faces $\{\bx_2^\ell,\by^\ell\}$. This allows employing our shared-representation regression technique for designing a predictor of $Y$ based on $X_1$. If, in addition, several subjects were photographed more than once, then we may construct a third set $\{\bx_1^\ell,\by^\ell\}$, containing pairs of images of smiling and neutral-expression faces. In this case, we can apply regression with side information, as discussed in Section~\ref{sec:RegSideInfo}.

\begin{figure*}\centering
\includegraphics[width=0.1\textwidth, trim=2.2cm 1.9cm 2.2cm 0.5cm, clip]{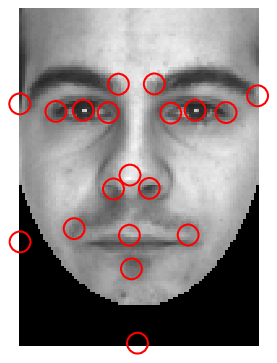}
\includegraphics[width=0.1\textwidth, trim=2.2cm 1.9cm 2.2cm 0.5cm, clip]{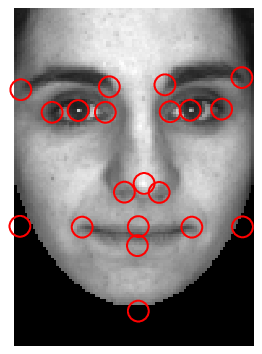}
\includegraphics[width=0.1\textwidth, trim=2.2cm 1.9cm 2.2cm 0.5cm, clip]{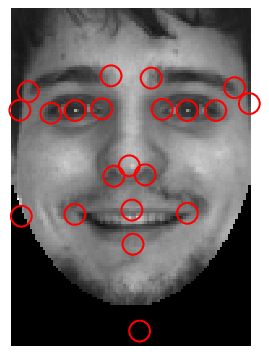}
\includegraphics[width=0.1\textwidth, trim=2.2cm 1.9cm 2.2cm 0.5cm, clip]{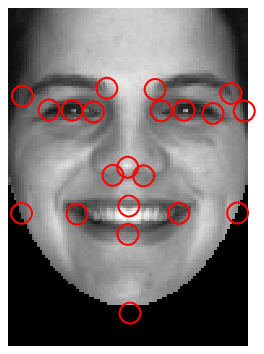}
\caption{Annotated images from the AR database.} \label{fig:database}
\end{figure*}

Figure~\ref{fig:database} depicts several manually annotated neutral and smiling facial images taken from the AR database \citep{AR98}. The point annotations were taken from \url{http://www-prima.inrialpes.fr/FGnet/data/05-ARFace/tarfd_markup.html}. The images were scaled, rotated and cropped into an ellipsoidal template such that the eyes appear at predefined locations. In practice, this can be performed automatically \citep{RY98,MICHAELI:2010}. To apply our methods, we normalized the images to be of zero mean and unity norm and reduced them to $86$ dimensions using PCA. The nonlinear regression scheme we used as a building block in our methods was first-order polynomial regression with a Gaussian kernel. The bandwidth of the kernel was adaptively tuned to be a constant times the root of the average squared distance between the query and the training data points.

\begin{figure*}\centering
\includegraphics[width=0.1\textwidth, trim=2.2cm 1.9cm 2.2cm 0.5cm, clip]{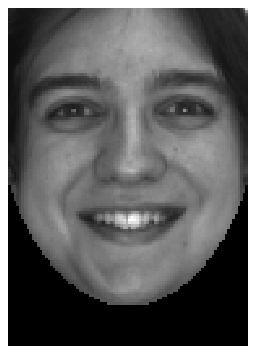} \includegraphics[width=0.1\textwidth, trim=2.2cm 1.9cm 2.2cm 0.5cm, clip]{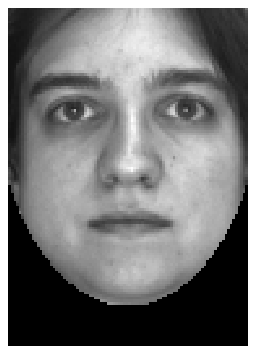} \includegraphics[width=0.1\textwidth, trim=2.2cm 1.9cm 2.2cm 0.5cm, clip]{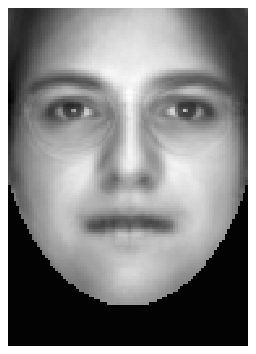}
\includegraphics[width=0.1\textwidth, trim=2.2cm 1.9cm 2.2cm 0.5cm, clip]{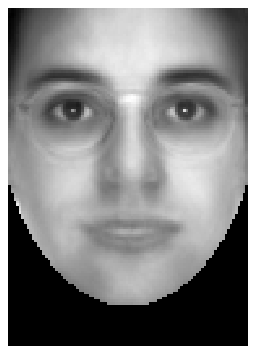} \includegraphics[width=0.1\textwidth, trim=2.2cm 1.9cm 2.2cm 0.5cm, clip]{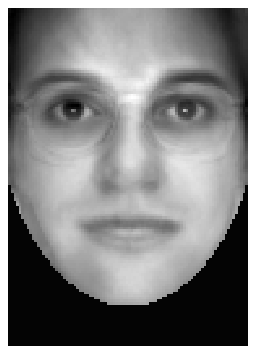}\\
\includegraphics[width=0.1\textwidth, trim=2.2cm 1.9cm 2.2cm 0.5cm, clip]{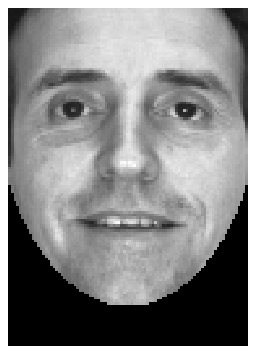} \includegraphics[width=0.1\textwidth, trim=2.2cm 1.9cm 2.2cm 0.5cm, clip]{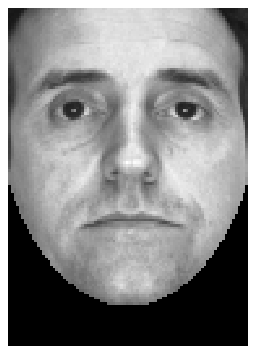} \includegraphics[width=0.1\textwidth, trim=2.2cm 1.9cm 2.2cm 0.5cm, clip]{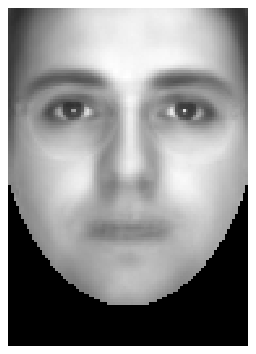}
\includegraphics[width=0.1\textwidth, trim=2.2cm 1.9cm 2.2cm 0.5cm, clip]{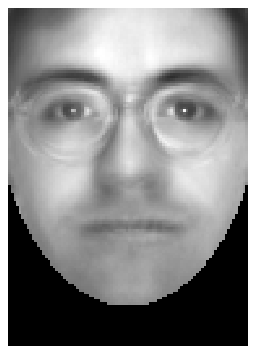} \includegraphics[width=0.1\textwidth, trim=2.2cm 1.9cm 2.2cm 0.5cm, clip]{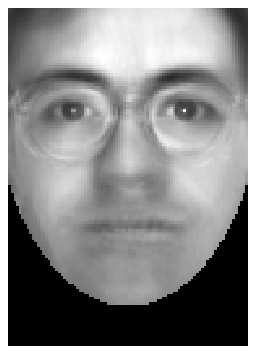}\\
\caption{Neutral expression synthesis from smiling images. From left to right: query, ground truth, direct nonlinear regression, shared-representation nonlinear regression (Section~\ref{sec:SharedRep}), linear regression with nonlinear side information (Section~\ref{sec:RegSideInfo}).}
\label{fig:FacesResults}
\end{figure*}

Figure~\ref{fig:FacesResults} demonstrates the results obtained with our approach in several settings. The two leftmost columns correspond to the query smiling face and the corresponding desired (unobserved) neutral expression image. The third column shows the result of directly performing regression using $118$ pairs of smile/neutral images. The fourth column is the result of performing shared representation regression via \eqref{eq:sharedRepEst}, using a training set of $38$ annotated smiling faces and a set of $40$ annotated neutral images (of different subjects). The rightmost column uses, in addition to these two sets, a training set comprising $40$ pairs of images of neutral and smiling expressions to perform linear regression with nonlinear side information (equation \eqref{eq:linRegNonlinSideInfo}).

Table~\ref{tbl:FaceResults} shows the root MSE (RMSE), $(\EE[\|Y-\hat{Y}\|^2])^{\frac{1}{2}}$, attained in each of the settings. As expected, using direct training with $118$ examples yields the best results (lowest RMSE). It can be seen that employing two sets with roughly $40$ examples each, instead of direct training, leads to an increase in the RMSE by $41\%$. This gap is reduced to $32\%$ with the aid of an additional set of $40$ direct training pairs. Perceptually, the images produced by the indirect methods do not seem to be much worse than those obtained with direct training. Note that the spatial smoothing apparent in all methods is due to the fact that any regression methods boils down at the end to some sort of averaging of many images from the training set. It is also important to note that the vague traces of glasses in the last two columns are no coincidence. Specifically, when there are no (or very few) joint examples of smile/neutral faces, no method can ever be able to determine whether the person wears glasses or not. This is because we only know how the smiling images (pixel values) relate to the geometry (point annotations) and how the geometry relates to the neutral images. Now, for every possible geometry, roughly half the people in the neutral database wear glasses and half not.

\begin{table}
\caption{Performance of Neutral Expression Synthesis Methods}
\label{tbl:FaceResults}
\centering
\begin{tabular}{|c|c|}
\hline
Setting & RMSE \\ \hline\hline
Direct nonlinear regression                       & 0.193 \\ \hline
Shared-representation nonlinear regression        & 0.263 \\ \hline
Linear regression with nonlinear side information & 0.247 \\ \hline
\end{tabular}
\end{table}

\subsection{Audio-Visual Word Recognition}
Although the entire discussion in this paper has focused on regression, similar methods can be developed for classification tasks. To support our claim, we now illustrate that this can even be achieved by using the naive approach of performing regression and then quantizing the output in order to obtain a classification rule.

Specifically, we now consider the tasks of spoken digit classification from audio-only and video-only measurements. To study this task, we used the Grid Corpus \citep{CBCS06}, which consists of speakers saying simple-structured sentences. Every sentence contains one digit, which we isolated using the supplied transcriptions. We constructed three distinct training sets: one of labeled audio examples ($4$ males, $4$ females), one of visual examples ($4$ males, $4$ females), and one of unlabeled audio-visual examples ($6$ males, $4$ females). Six speakers were used for testing ($3$ males, $3$ females).

\begin{figure*}\centering
\includegraphics[scale=0.5, trim=2cm 0.5cm 2cm 1cm, clip]{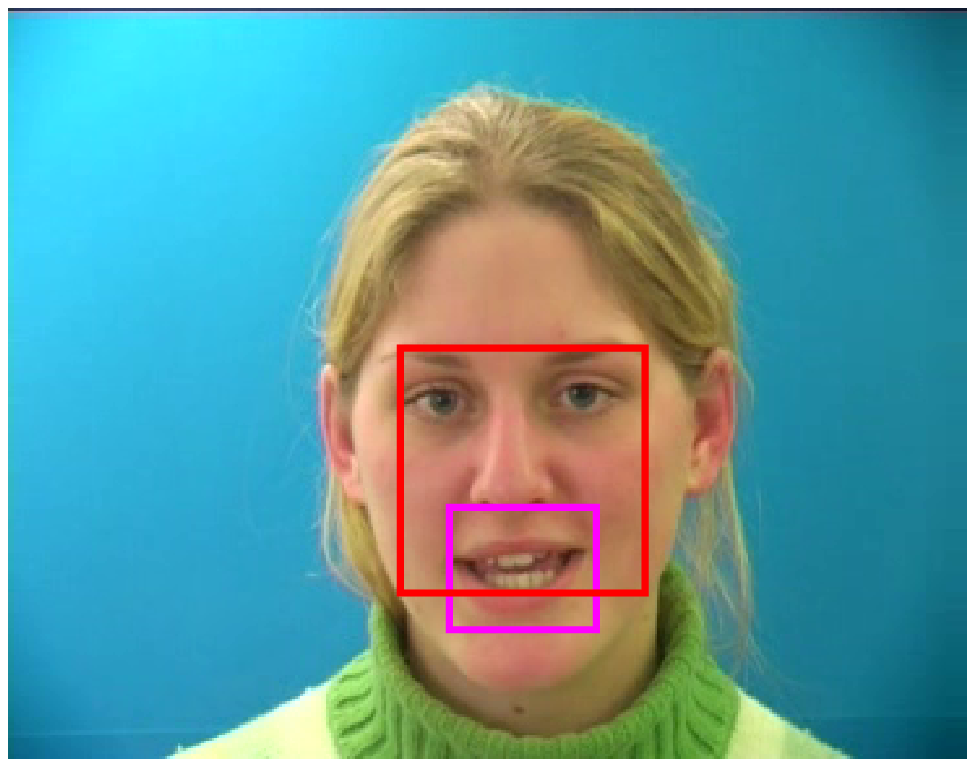}
\includegraphics[scale=1, trim=2cm 1cm 2cm 1cm, clip]{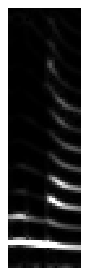}
 \includegraphics[scale=1, trim=2cm 1cm 2cm 1cm, clip]{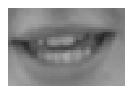}
\caption{Processing of the video and audio of a speaker saying the word `nine'. From left to right: lip detection, spectogram, extracted lip region.} \label{fig:Speaker}
\end{figure*}

To process the video, we converted the images to gray scale, used the face detection method of \cite{KBFS04}, and then applied several mean-shift iterations on the gradient image map in order to extract the lip region in the first image of each frame-bunch. Segments of duration $320$msec were used for recognition. This corresponded to $8$ consecutive video frames (at a rate of $25$ frames per second) and $1600$ audio samples (at a sampling rate of $5$KHz). The image frames were reduced to $10$ dimensions using PCA, resulting in an $80$-dimensional video feature-vector. The processing of the audio was performed by computing spectograms with windows of duration $10$msec and an overlap of $2.5$msec. The dimension of the spectogram was reduced to $180$ to constitute the audio features. In all experiments $Y$ was a $10$-dimensional vector with $1$ at the location corresponding to the spoken digit and $0$ elsewhere. Figure~\ref{fig:Speaker} visualizes the basic audio-visual preprocessing.

As mentioned above, our approach is designed for regression, so that the predicted $\hat{Y}$ is a continuous variable. To perform classification, we chose the maximal element in $\hat{Y}$. For simplicity, $\A$ and $\B$ were taken as the sets of all linear functions (linear regression). This choice yields rather poor classification results based solely on audio or solely on video. Our goal, though, is to demonstrate that even with such naive single-domain predictors, we can attain good recognition accuracy by using our approach, which cleverly fuses the two domains.

Table~\ref{tbl:results} shows the accuracy of the our approach and for reference also presents the results obtained with the deep restricted Boltzmann machine (RBM) of \cite{NKKNLN10} on the CUAVE dataset \citep{PGTG93}. The Grid corpus used here is more challenging in that the digits appear within sentences, rather than individually. As can be seen, the single-domain predictors we start with perform relatively poorly (rows $1$ and $2$).
Nevertheless, in the shared-representation settings (rows $3$ and $4$), our predictors perform much better than the RBM method, even for a harder dataset. Their accuracy is only between $7\%$ and $20\%$ worse than the corresponding single domain estimators (rows $1$ and $2$, respectively). By contrast, the difference in success rates for the RBM predictor is between $30\%$ and $70\%$.

\begin{table}
\caption{Audio-Visual Digit Classification Performance}
\label{tbl:results}
\centering
\begin{tabular}{|c|c|c|c|}
\hline
\multicolumn{2}{|c|}{Features}                                    & \multicolumn{2}{|c|}{Accuracy} \\
\multirow{2}{*}{Training}       & \multirow{2}{*}{Testing}        & Minimax       & Deep RBM    \\
                                &                                 & {\footnotesize(Grid corpus)} & {\footnotesize(CUAVE)}\\ \hline\hline
Audio          & Audio          & 69.3\%                          &  95.8\%      \\ \hline
Video          & Video          & 52.0\%                          &  69.7\%      \\ \hline
Video          & Audio          & 50.1\%                          &  27.5\%      \\ \hline
Audio          & Video          & 44.6\%                          &  29.4\%      \\ \hline
\end{tabular}
\end{table}

\section{Conclusion}
In this paper, we analyzed the problems of multi-domain and single-domain regression in settings involving distinct unpaired labeled training sets for the different domains and a large unlabeled set of paired examples from all domains. We derived minimax-optimal results and obtained closed form solutions for many practical scenarios. We used the resulting expressions to study when training data from a domain, which is not available during testing, can help. In particular, we showed that in the setting of cross-modality learning, originally presented in \cite{NKKNLN10}, there is no advantage in using the training data from the unobserved domain, at least from a worst-case perspective. We demonstrated our methods in the context of synthesis of a neutral expression face from an image of a smiling subject and in the context of audio-visual spoken digit recognition. In the latter setting, we demonstrated that our approach may be more effective than that proposed in \cite{NKKNLN10}. This is despite the fact that our method is designed for regression rather than classification and even though we applied it on a more challenging audio-visual sentence corpus.

\appendix


\section{Proof of Theorem~\ref{thm:MxMultiview}}
\label{sec:ProofThmMxMultiview}

We begin by proving claim \ref{ItemRhoStarInvariant}.
Since $\A$ is a linear subspace, the orthogonality principle implies that $\varphi_{\A}(X_1)$ is the unique estimator satisfying
\begin{equation}
\EE\left[(Y-\varphi_{\A}(X_1))^T\varphi(X_1)\right]=0
\end{equation}
for every $\varphi\in\A$. Consequently, for every $\varphi\in\A$ we have that
\begin{equation}\label{eq:Xphi}
\EE\left[Y^T\varphi(X_1)\right]=\EE\left[\varphi_{\A}(X_1)^T\varphi(X_1)\right].
\end{equation}
Similarly, for every $\psi\in\B$ we have that
\begin{equation}\label{eq:Xpsi}
\EE\left[Y^T\psi(X_2)\right]=\EE\left[\psi_{\B}(X_2)^T\psi(X_2)\right].
\end{equation}
Finally, as $\C=\A+\B$, the set $\C$ is a subspace as well. Therefore, $\rho_{\C}$ of \eqref{eq:RhoStarMinimumYZ} is the unique estimator satisfying
\begin{align}\label{eq:Xrho}
\EE\left[Y^T(\varphi(X_1)+\psi(X_2))\right]
=\EE\left[\rho_{\C}(X_1,X_2)^T(\varphi(X_1)+\psi(X_2))\right]
\end{align}
for every $\varphi\in\A$ and $\psi\in\B$. Substituting \eqref{eq:Xphi} and \eqref{eq:Xpsi}, condition \eqref{eq:Xrho} reduces to the requirement that
\begin{align}\label{eq:rhoStarOrtCond}
\EE\left[\varphi_{\A}(X_1)^T\varphi(X_1)\right]+\EE\left[\psi_{\B}(X_2)^T\psi(X_2)\right] =\EE\left[\rho_{\C}(X_1,X_2)^T(\varphi(X_1)+\psi(X_2))\right]
\end{align}
for every $\varphi\in\A$ and $\psi\in\B$. Now, the $\A$- and $\B$-optimal estimators of $Y$ from $X_1$ and $X_2$ are fixed over $\F$ (given by $\varphi_{\A}$ and $\psi_{\B}$, respectively). Furthermore, all expectations in \eqref{eq:rhoStarOrtCond} are with respect to $F_{X_1X_2}$, which is also fixed over $\F$. This implies that the function $\rho_{\C}$ does not depend on the choice of $F_{X_1X_2Y}\in\F$, completing the proof of claim \ref{ItemRhoStarInvariant}.

To prove claim \ref{ItemMSEInvariant}, we note that from the orthogonality principle \eqref{eq:Xrho} follows the Pythagorean relation
\begin{align}\label{eq:XminusRhoStar}
\EE\left[\|Y-\rho_{\C}(X_1,X_2)\|^2\right] = \EE\left[\|Y\|^2\right] - \EE\left[\|\rho_{\C}(X_1,X_2)\|^2\right].
\end{align}
The first term on the right-hand side equals $c$ for every $F_{X_1X_2Y}\in\F$. We have also seen that $\rho_{\C}(X_1,X_2)$ is fixed over $\F$. Moreover, the expectation in the second term is with respect to $\F_{X_1X_2}$, which is fixed over $\F$. Therefore, the second term, as well, does not depend on the choice of $F_{X_1X_2Y}\in\F$. This completes the proof of claim \ref{ItemMSEInvariant}.

Lastly, we prove claim \ref{ItemRhoStarMinimax}. To do so, we first note that $\varphi_{\A}(X_1)$ and $\psi_{\B}(X_2)$ are not only the $\A$- and $\B$-optimal estimators of $Y$ based on $X_1$ and $X_2$, respectively; they are also the $\A$- and $\B$-optimal estimators of $\rho_{\C}(X_1,X_2)$. To see this, note that both $\A$ and $\B$ are contained in $\C$. Consequently, the orthogonality principle implies that for every $\varphi\in\A$ (which is also in $\C$), we have
\begin{align}
\EE[\|Y-\varphi(X_1)\|^2] &= \EE[\|Y-\rho_{\C}(X_1,X_2)\|^2] +\EE[\|\rho_{\C}(X_1,X_2)-\varphi(X_1)\|^2].
\end{align}
As the first term does not depend on $\varphi$, we see that minimization of the MSE over $\varphi\in\A$ is equivalent to minimization of the second term alone. Thus, $\varphi_{\A}(X_1)$ is the $\A$-optimal estimate of $\rho_{\C}(X_1,X_2)$ given $X_1$. The same argument can be invoked to deduce that $\psi_{\B}(X_2)$ is the $\B$-optimal estimate of $\rho_{\C}(X_1,X_2)$ from $X_2$.

A second observation we need for proving claim \ref{ItemRhoStarMinimax} follows from the fact that $\A$ and $\B$ are linear subspaces. Specifically, this implies that if $\varphi_1^*(V)$ and $\varphi_2^*(V)$ are the $\A$-optimal estimates of the two RVs $W_1$ and $W_2$, respectively, based on the RV $V$, then the $\A$-optimal estimate of $W_1+W_2$ is $\varphi_1^*(V)+\varphi_2^*(V)$. This can be seen by noting that the estimator $\varphi_1^*(V)+\varphi_2^*(V)$ satisfies the orthogonality principle, namely for any $\varphi\in\A$ we have that
\begin{align}
\EE[(W_1+W_2-\varphi_1^*(W_1)-\varphi_2^*(W_1))^T\varphi(W_1)] &=\EE[(W_1-\varphi_1^*(W_1))^T\varphi(W_1)] + \EE[(W_2-\varphi_2^*(W_1))^T\varphi(W_1)] \nonumber\\
&= 0.
\end{align}
The statement also holds, of course, with respect to $\B$-optimal estimates.

Following these two observations, for any $F_{X_1X_2Y}\in\F$, setting $\tilde{Y}=2\rho_{\C}(X_1,X_2)-Y$ results in a distribution $F_{X_1X_2\tilde{Y}}$ that also belongs to $\F$. This is because the $\A$-optimal estimate of $\tilde{Y}$ from $X_1$ equals twice the $\A$-optimal estimate of $\rho_{\C}(X_1,X_2)$ from $X_1$ (which is $\varphi_{\A}(X_1)$) minus the $\A$-optimal estimate of $Y$ from $X_1$ (which is also $\varphi_{\A}(X_1)$). Namely, the $\A$-optimal estimate of $\tilde{Y}$ from $X_1$ is $\varphi_{\A}(X_1)$. Similarly, the $\B$-optimal estimate of $\tilde{Y}$ from $X_2$ is $\psi_{\B}(X_2)$. Finally, due to the orthogonality principle, the second-order moment of $\tilde{Y}$ is given by
\begin{align}
\EE[\|\tilde{Y}\|^2] &= \EE[\|\rho_{\C}(X_1,X_2)\|^2] + \EE[\|Y-\rho_{\C}(X_1,X_2)\|^2] \nonumber\\
&= \EE[\|\rho_{\C}(X_1,X_2)\|^2] + \EE[\|Y\|^2] - \EE[\rho_{\C}(X_1,X_2)\|^2] \nonumber\\
&= c.
\end{align}

We now use this fact to prove claim \ref{ItemRhoStarMinimax}. The orthogonality principle \eqref{eq:Xrho} implies that the MSE attained by any estimator $\rho$ satisfies
\begin{align}\label{eq:minimaxMSEProof}
\EE\left[\|Y-\rho(X_1,X_2)\|^2\right] &=\EE\left[\|Y-\rho_{\C}(X_1,X_2)\|^2\right]
+\EE\left[\|\rho_{\C}(X_1,X_2)-\rho(X_1,X_2)\|^2\right] \nonumber\\
&\hspace{4cm}+2\EE\left[(Y-\rho_{\C}(X_1,X_2))^T (\rho_{\C}(X_1,X_2)-\rho(X_1,X_2))\right] \nonumber\\
&= \EE\left[\|Y-\rho_{\C}(X_1,X_2)\|^2\right] +\EE\left[\|\rho_{\C}(X_1,X_2)-\rho(X_1,X_2)\|^2\right] \nonumber\\
&\hspace{4cm}+2\EE\left[(\rho_{\C}(X_1,X_2)-Y)^T \rho(X_1,X_2)\right].
\end{align}
The first term in this expression is not a function of $\rho$ and, as we have seen in \eqref{eq:XminusRhoStar}, is constant as a function of $F_{X_1X_2Y}$ over $\F$. The second term is a function of $\rho$, but since the expectation is with respect to $F_{X_1X_2}$, it is constant as a function of $F_{X_1X_2Y}$ over $\F$. Therefore,
\begin{align}
\min_{\rho}\sup_{F_{X_1X_2Y}\in\F}{\rm MSE}(F_{X_1X_2Y},\rho)
= \EE\!\left[\|Y-\rho_{\C}(X_1,X_2)\|^2\right] &+ \min_{\rho}\!\Big\{\EE\!\left[\|\rho_{\C}(X_1,X_2)-\rho(X_1,X_2)\|^2\right] \nonumber\\
&+ \sup_{F_{X_1X_2Y}\in\F} 2\EE\!\left[(\rho_{\C}(X_1,X_2)-Y)^T \rho(X_1,X_2)\right]\Big\}.
\end{align}
We saw that for every $F_{X_1X_2Y}\in\F$ setting $\tilde{Y}=2\rho_{\C}(X_1,X_2)-Y$ results in a distribution $F_{X_1X_2\tilde{Y}}$ that also belongs to $\F$. Now, with $F_{X_1X_2\tilde{Y}}$, the expression $2\EE[(\rho_{\C}(X_1,X_2)-\tilde{Y})^T \rho(X_1,X_2)]$ equals $-2\EE[(\rho_{\C}(X_1,X_2)-Y)^T \rho(X_1,X_2)]$. Consequently, the maximum of this term over $F_{X_1X_2Y}\in\F$ is necessarily nonnegative. We thus have that
\begin{align}\label{eq:minimaxMSEProof2}
\min_{\rho}\sup_{F_{X_1X_2Y}\in\F}{\rm MSE}(F_{X_1X_2Y},\rho) &\geq \EE\left[\|Y-\rho_{\C}(X_1,X_2)\|^2\right] +\min_{\rho}\EE\!\left[\|\rho_{\C}(X_1,X_2)-\rho(X_1,X_2)\|^2\right] \nonumber\\
&= \EE\left[\|Y-\rho_{\C}(X_1,X_2)\|^2\right],
\end{align}
where we used the fact that the minimal value of $0$ is attained with $\rho(X_1,X_2)=\rho_{\C}(X_1,X_2)$.

We have established a lower bound on the worst-case MSE of any estimator. Next, we show that the estimator $\rho(X_1,X_2)=\rho_{\C}(X_1,X_2)$ attains this bound, which proves that it is minimax-optimal. Indeed, substituting this solution into \eqref{eq:minimaxMSEProof}, we find that
\begin{align}
\sup_{F_{X_1X_2Y}\in\F}{\rm MSE}(F_{X_1X_2Y},\rho_{\C}) = \EE\!\left[\|Y-\rho_{\C}(X_1,X_2)\|^2\right],
\end{align}
completing the proof.


\section{Proof of Theorems~\ref{thm:MxRegMultiview} and \ref{thm:MxRegSingleview}}
\label{sec:ProofThmMxReg}
We simultaneously prove Theorems~\ref{thm:MxRegMultiview} and \ref{thm:MxRegSingleview} by using an auxiliary RV $Z$, which can be any (fixed) function of $X_1$ and $X_2$. Therewith, we will study the solution to
\begin{equation}
\argmin_{\rho}\sup_{F_{X_1X_2Y}\in\F}{\rm REG}(F_{X_1X_2Y},\rho),
\end{equation}
where minimization is performed over all functions $\rho$ of $Z$ and the regret is with respect to $\EE[Y|Z]$. Specifically, we will show that the solution to this problem is given by $\EE[\rho_{\rm M}(X_1,X_2)|Z]$. Setting, $Z=(X_1^T,X_2^T)^T$, we get $\EE[\rho_{\rm M}(X_1,X_2)|Z] = \rho_{\rm M}(X_1,X_2)$, proving Theorem~\ref{thm:MxRegMultiview}. Setting $Z=X_1$, the solution becomes $\EE[\rho_{\rm M}(X_1,X_2)|X_1]$, proving Theorem~\ref{thm:MxRegSingleview}.

Expressing $Y=\rho_{\rm M}(X_1,X_2)+(Y-\rho_{\rm M}(X_1,X_2))$, the regret of any estimator $\rho(Z)$ can be written as
\begin{align}
\EE\left[\|\EE[Y|Z]-\rho(Z)\|^2\right] &= \EE\left[\|\EE[\rho_{\rm M}(X_1,X_2)|Z]-\rho(Z)\|^2\right] + \EE\left[\|\EE[Y- \rho_{\rm M}(X_1,X_2)|Z]\|^2\right] \nonumber\\
&\hspace{0.5cm} + 2\EE\left[\EE[Y-\rho_{\rm M}(X_1,X_2)|Z]^T (\EE[\rho_{\rm M}(X_1,X_2)|Z]-\rho(Z))\right].
\end{align}
Since the marginal distribution $F_{X_1X_2}$ is fixed over $\F$, the first term in the above expression does not depend on the choice of $F_{X_1X_2Y}\in\F$. Consequently,
\begin{align}\label{eq:supRegXi}
\sup_{F_{X_1X_2Y}\in\F}{\rm REG}(F_{X_1X_2Y},\rho) &= \EE[\|\EE[\rho_{\rm M}(X_1,X_2)|Z]-\rho(Z)\|^2] + \sup_{F_{X_1X_2Y}\in\F} \Bigg\{\EE[\|\EE[Y-\rho_{\rm M}(X_1,X_2)|Z]\|^2] \nonumber\\
&\hspace{2cm} +2\EE\Big[\EE[Y-\rho_{\rm M}(X_1,X_2)|Z]^T (\EE[\rho_{\rm M}(X_1,X_2)|Z]-\rho(Z))\Big]\Bigg\}.
\end{align}
As we have seen in Appendix~\ref{sec:ProofThmMxMultiview}, for every $F_{X_1X_2Y}\in\F$ setting $\tilde{Y}=2\rho_{\rm M}(X_1,X_2)-Y$ results in a distribution $F_{X_1X_2\tilde{Y}}$ that also belongs to $\F$. Now, $\tilde{Y}-\rho_{\rm M}(X_1,X_2)=-(Y-\rho_{\rm M}(X_1,X_2))$, implying that if $F_{X_1X_2Y}$ maximizes the first term within the braces, then either $F_{X_1X_2Y}$ or $F_{X_1X_2\tilde{Y}}$ yields at least the same value for the objective comprising both terms. Therefore,
\begin{align}
\min_{\rho}\sup_{F_{X_1X_2Y}\in\F}{\rm REG}(F_{X_1X_2Y},\rho) &\geq \min_{\rho}\EE\left[\|\EE[\rho_{\rm M}(X_1,X_2)|Z]-\rho(Z)\|^2\right] \nonumber\\
&\hspace{3cm}+ \sup_{F_{X_1X_2Y}\in\F} \EE\left[\|\EE[Y-\rho_{\rm M}(X_1,X_2)|Z]\|^2\right] \nonumber\\
&= \sup_{F_{X_1X_2Y}\in\F} \EE\left[\|\EE[Y-\rho_{\rm M}(X_1,X_2)|Z]\|^2\right],
\end{align}
where the last equality is due to the fact that $\rho(Z)=\EE[\rho_{\rm M}(X_1,X_2)|Z]$ achieves the minimal value of $0$ in the first term.

We established a lower bound on the worst-case regret of any estimator. Next, we show that the estimator $\rho^*(Z)=\EE[\rho_{\rm M}(X_1,X_2)|Z]$ attains this bound, which proves that it is minimax-optimal. Indeed, substituting this solution into \eqref{eq:supRegXi}, we find that
\begin{align}
\sup_{F_{X_1X_2Y}\in\F} {\rm REG}(F_{X_1X_2Y}, \rho_{\rm M}) = \sup_{F_{X_1X_2Y}\in\F}\EE\left[\|\EE[Y-\rho_{\rm M}(X_1,X_2)|Z]\|^2\right],
\end{align}
completing the proof.


\section{Proof of Theorem~\ref{thm:SemiParametric}}
\label{sec:ProofSemiParametric}
To prove the claim, we show that the estimation error corresponding to $\rho_{\C}(X_1,X_2)$ of \eqref{eq:SemiParametric} is uncorrelated with every RV of the form $\varphi(X_1)+\psi(X_2)$ with $\varphi\in\A$ and $\psi\in\B$. Indeed, for every $\varphi\in\A$, the estimator $\rho_{\C}(X_1,X_2)$ of \eqref{eq:SemiParametric} satisfies
\begin{align}
\EE\left[(Y - \rho_{\C}(X_1,X_2))^T \varphi(X_1)\right] &= \EE\left[(Y - \varphi^*(X_1))^T \varphi(X_1)\right] - \EE\left[\rho^T_{\A,\B}(X_1,X_2) \varphi(X_1)\right] \nonumber\\
&= \EE\left[\left(\psi(X_2)-\eta_{\psi}(X_1)\right)^T \varphi(X_1)\right] \nonumber\\
&= 0,
\end{align}
where we used the orthogonality principle. To prove orthogonality with respect to RVs of the form $\psi(X_2)$, with $\psi\in\B$, we write $\psi(X_2)=\psi(X_2)-\eta_{\psi}(X_1)+\eta_{\psi}(X_1)$, where $\eta_{\psi}(X_1)$ is the $\A$-optimal estimate of $\psi(X_2)$ based on $X_1$. By the orthogonality principle, the errors $Y - \varphi_{\A}(X_1)$ and $\rho_{\A,\B}(X_1,X_2)=\psi(X_2)-\eta_{\psi}(X_1)$ are uncorrelated with any RV $\eta(X_1)$, where $\eta\in\A$, and thus in particular with the term $\eta_{\psi}(X_1)$. Therefore, we have that
\begin{align}
\EE\left[(Y - \hat{Y})^T \psi(X_2)\right] &= \EE\left[\left(Y - \varphi_{\A}(X_1) - \rho_{\A,\B}(X_1,X_2)\right)^T \left(\psi(X_2)-\eta_{\psi}(X_1)\right)\right] \nonumber\\
&= \EE\left[\left(Y - \rho_{\A,\B}(X_1,X_2)\right)^T \left(\psi(X_2)-\eta_{\psi}(X_1)\right)\right] \nonumber\\
&=0.
\end{align}
Here, the second equality results from the fact that the term $\psi(X_2)-\eta_{\psi}(X_1)$ is orthogonal to every RV $\varphi(X_1)$, where $\varphi\in\A$ and, in particular, to $\varphi_{\A}(X_1)$. The third equality follows from the fact that $\rho_{\A,\B}(X_1,X_2)$ is the MMSE estimate of $Y$ among all functions of the form $\psi(X_2)-\eta_{\psi}(X_1)$, with $\psi$ being some function in $\B$ and $\eta_{\psi}(X_1)$ being the $\A$-optimal estimator of $\psi(X_2)$ from $X_1$. Consequently, the error $Y-\rho_{\A,\B}(X_1,X_2)$ is orthogonal to every RV of the form $\psi(X_2)-\eta_{\psi}(X_1)$, and, in particular, to $\psi(X_2)-\eta_{\psi}(X_1)$.

\section*{Funding}
This work was supported in part by the Office of Naval Research; Army Research Office; Defense Advanced Research Projects Agency; National Security Science and Engineering Faculty Fellowship; National Science Foundation; Israel Science Foundation [grant number~170/10]; Ollendorff Minerva Foundation; and Google Research Award.

\bibliographystyle{imaiai}
\bibliography{IEEEabrv,SemiSupervisedPartialKnowledge}

\end{document}